\documentclass{article}

 \usepackage[main, final]{neurips_2025}

\usepackage[utf8]{inputenc} % allow utf-8 input
\usepackage[T1]{fontenc}    % use 8-bit T1 fonts
\usepackage{hyperref}       % hyperlinks
\usepackage{url}            % simple URL typesetting
\usepackage{booktabs}       % professional-quality tables
\usepackage{amsfonts}       % blackboard math symbols
\usepackage{nicefrac}       % compact symbols for 1/2, etc.
\usepackage{microtype}      % microtypography
\usepackage{xcolor}         % colors
\usepackage{multirow}
\usepackage{enumitem}
\usepackage[table]{xcolor}
\usepackage[most]{tcolorbox}
\usepackage{minitoc}
\usepackage{pifont}% http://ctan.org/pkg/pifont
\newcommand{\cmark}{\ding{51}}%
\newcommand{\xmark}{\ding{55}}%

\title{OmniSpace: Efficient Geometry Awareness for Autonomous Vehicles MLLMs}
\author{%
Hao Vo$^{1*}$,
Phu Loc Nguyen$^{1*}$,
Khoa Vo$^{1}$,
Sieu Tran$^{1}$,
Duc Minh Nguyen$^{1}$,
Ngo Xuan Cuong$^{1}$, \\
\bf Nghi D. Q. Bui$^{2}$, 
 Anh Nguyen$^{3}$,
Duy Minh Ho Nguyen$^{4}$,
Ngan Le$^{1}$\\
$^{1}$University of Arkansas, USA \quad
$^{2}$Google Research, Google \quad \\
$^{3}$University of Liverpool, UK \quad
$^{4}$Max Planck Research School for Intelligent Systems \quad \\
$^{*}$Equal contribution \\
\vspace{-6mm}
}

\usepackage{graphicx}
\usepackage{amsmath} 
\usepackage{amssymb}
\usepackage{xspace}
\usepackage{wrapfig}
\usepackage{graphicx}
\usepackage{color}
\usepackage{soul}
\usepackage{pifont}
\newcommand*{\rttensortwo}[1]{\bar{\bar{#1}}}
\newcommand{\model}{OmniSpace\xspace}

\definecolor{lightgraybg}{gray}{0.95}
\newcommand{\ourmodule}{OmniSpace\xspace}
\definecolor{qwenblue}{HTML}{E2D8F3}
\begin{document}
\doparttoc
\faketableofcontents

\maketitle

\begin{abstract}
Multimodal Large Language Models (MLLMs) have achieved remarkable performance on 2D visual tasks, yet enhancing their spatial intelligence for real-world applications such as Autonomous Vehicles (AV) remains an open challenge. Existing geometry-aware MLLMs typically rely on auxiliary 3D models at inference time, introducing pipeline complexity and the risk of cascading failures. In this paper, we present \textbf{\model}, a backbone-agnostic framework for geometry-aware spatial reasoning from purely camera-only observations. Motivated by our finding that current MLLMs are bottlenecked by weak cross-view correspondence and depth estimation, \textbf{\model} introduces a Camera Pose Injector, a Multi-view Epipolar Attention module, and a 3D Geometric Distillation objective that jointly address these two limitations by transferring geometric knowledge into the model. Extensive experiments show that \textbf{\model} surpasses existing methods on planning benchmarks (nuScenes, Bench2Drive), risk detection (nuInstruct), language (Omnidrive), and generalization (DriveBench).
\end{abstract}

\section{Introduction}

% \begin{wrapfigure}{r}{0.38\textwidth}
%    \vspace{-12pt}
%    \includegraphics[width=0.38\textwidth]{images/method_comparision_v8.pdf}
%    \vspace{-15pt}
%    \caption{Comparison between existing paradigms and \textbf{\model}.}
%    \label{fig:method_comparison}
%    \vspace{-15pt}
% \end{wrapfigure}
Rapid advancements and impressive performance of Multimodal Large Language Models (MLLMs) \cite{wang2024qwen2,bai2025qwen3,hurst2024gpt,li2024llava,lin2024vila,liu2023visual,team2024gemini,wang2025internvl3,wu2024deepseek,vo2024henasy,truong2025directed} have positioned them as promising, generalizable alternatives to traditional Autonomous Vehicles (AV) systems \cite{hu2023planning,jiang2023vad,weng2024drive,vo2026semlt3d}. A broad spectrum of MLLM-based methods has emerged, ranging from multi-stage pipelines~\cite{li2025spacedrive, tian2024drivevlm} to end-to-end frameworks~\cite{zeng2025futuresightdrive,xu2024vlm}, from static-frame methods \cite{wang2026vggdrive,gholami2025spatial} to dynamic models \cite{wang2025omnidrive, huang2025robotron, zeng2025futuresightdrive}, and from single-view approaches \cite{xu2024drivegpt4,xu2024vlm,hwang2024emma,tian2024drivevlm} to multi-view paradigms \cite{wang2025omnidrive,huang2025robotron}.

\begin{figure}
    \centering
    \includegraphics[width=\linewidth]{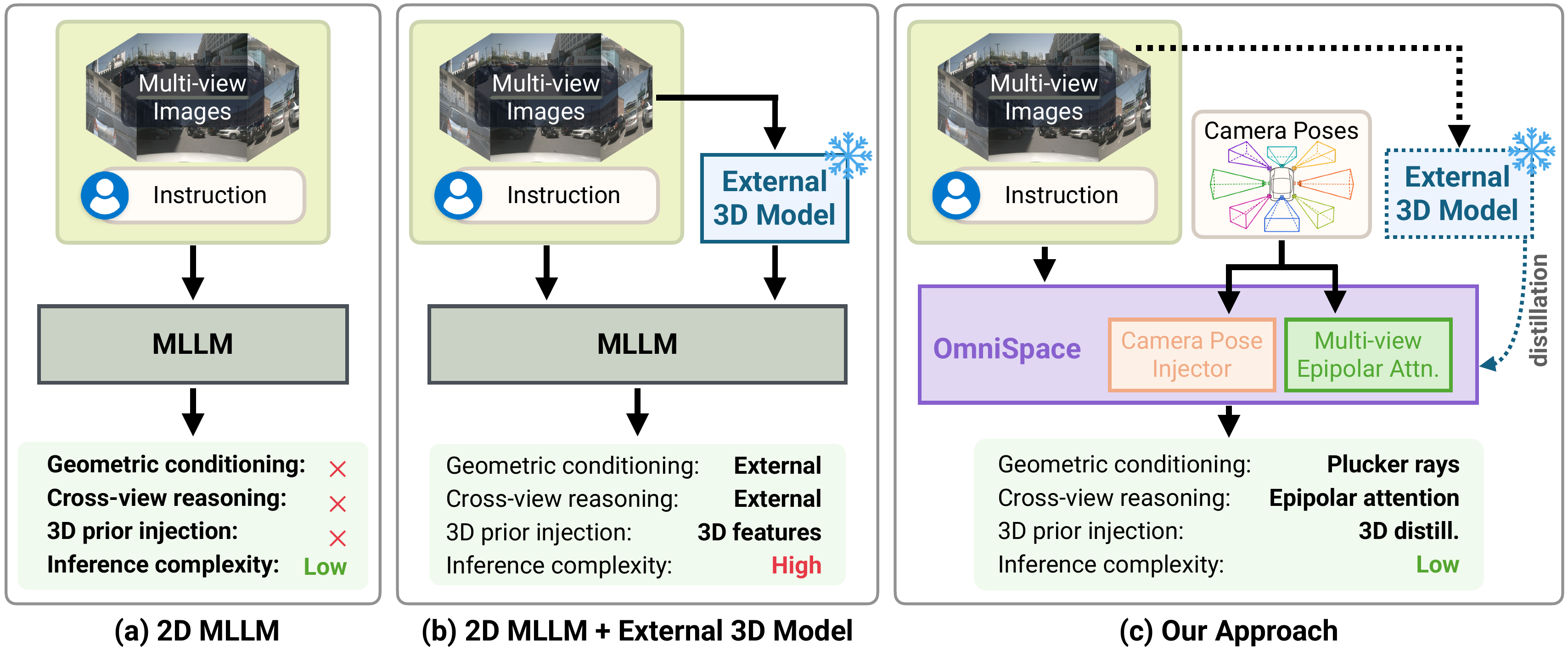}
    \caption{Comparison between existing paradigms and \textbf{\model}. }
    \label{fig:method_comparison}
\end{figure}

Despite these advances, many existing methods still formulate AV tasks as a conventional 2D visual autoregressive problem~\cite{huang2025robotron,zeng2025futuresightdrive,jiang2024senna} (Figure~\ref{fig:method_comparison}a). While this formulation offers a convenient path to adapting general-purpose MLLMs, it does so at the cost of geometric transparency: depth, scale, camera pose, and cross-view correspondence remain buried within image tokens, rather than being surfaced through geometry-aware visual features. This implicit treatment of geometry is particularly problematic in AV settings, where faithful geometry understanding lies at the heart of every downstream decision -- from collision avoidance to lane-level planning. A separate line of work addresses this limitation by introducing additional geometry-aware modules into MLLMs~\cite{wang2026vggdrive, luo2026last, wang2025omnidrive} (Figure~\ref{fig:method_comparison}b). 
%For example, OmniDrive~\cite{wang2025omnidrive} introduces Omni-Q and Omni-L, built upon StreamPETR~\cite{wang2023exploring}, to extract object-level embeddings for the language model, while VGGDrive~\cite{wang2026vggdrive} employs VGGT~\cite{wang2025vggt} as an auxiliary backbone and transfers its predictions to the MLLM through cross-attention. 
Although effective, this paradigm relies on test-time external 3D geometric model, which increases computational cost, complicates deployment, and makes the final prediction dependent on the quality of intermediate geometric estimates. This raises a natural question: \emph{Can MLLMs acquire geometry awareness directly, without sacrificing efficiency or deployment simplicity?}

To answer this question, we conduct a diagnostic investigation of current MLLMs in AV scene understanding. Drawing on principles from multi-view geometry~\cite{hartley2003multiple}, recovering a complete 3D scene requires both estimating per-image depth and reasoning about the geometric relationships among views. We therefore isolate two capabilities most directly tied to 3D scene understanding: (i)  \emph{depth estimation}, the ability to infer metric scene structure from camera observations; and (ii) \emph{cross-view correspondence}, the ability to associate the same scene elements across different camera views. In Section~\ref{sec:investigate}, we construct a benchmark to evaluate these two capabilities in four representative MLLMs~\cite{bai2025qwen3, wang2025internvl3, yao2024minicpm, team2024gemma} and examine whether stronger performance on these geometric diagnostics translates into stronger performance on AV-oriented tasks~\cite{wang2025omnidrive, caesar2020nuscenes, hu2023planning}. The investigation reveals a consistent trend: \textit{models with stronger depth estimation and cross-view correspondence also achieve better downstream AV tasks}. This finding suggests that these two geometric capabilities are key bottlenecks for MLLM-based AV understanding and motivates us to strengthen them directly within the MLLM, without adding external geometry modules at test time.

In this paper, we propose \textbf{\model} (Figure~\ref{fig:method_comparison}c), a new geometry-aware adaptation framework that strengthens cross-view correspondence and depth perception directly within MLLMs, without introducing additional test-time components. \model consists of two key designs: (i) To capture cross-view correspondence, we first introduce a \textit{Camera Pose Injector} that encodes each visual feature with its corresponding Pl\"ucker ray representation, providing explicit view-aware cues that inform the model of each pixel's 3D origin and viewing direction. Then, we propose \textit{Multi-view Epipolar Attention}, which uses epipolar geometry to restrict cross-view attention to geometrically plausible regions, thereby encouraging the model to associate corresponding scene elements across camera views. (ii) To equip the model with depth understanding, we introduce \textit{3D Geometric Distillation}, using a 3D foundation model~\cite{wang2025vggt} as a training-time teacher to distill geometric representations into \model, improving 3D scene understanding without incurring additional inference cost. Through these designs, \model directly enhances the geometric representations of MLLMs for AV scene understanding while preserving inference efficiency and deployment simplicity.

%First, we introduce a \textit{Camera Pose Injector} that encodes each visual feature with its corresponding Pl"{u}cker ray representation, providing explicit per-pixel ray and camera pose cues that ease depth estimation from multi-view inputs. Second, we propose \textit{Multi-view Epipolar Attention}, which uses epipolar geometry to restrict cross-view attention to geometrically plausible regions, thereby encouraging the model to associate corresponding scene elements across camera views. Third, we introduce \textit{3D Geometric Distillation}, which uses a 3D foundation model~\cite{wang2025vggt, wang2025pi} as a training-time teacher to supervise the geometric representation learned by \model, improving its 3D scene understanding without incurring additional inference cost. Through these designs, \model directly enhances the geometric representations of MLLMs for AV scene understanding while preserving inference efficiency and deployment simplicity.

Our key contributions are: (i) We identify depth estimation and cross-view correspondence as \textbf{two key geometric bottlenecks} for MLLM-based AV scene understanding, and validate their correlation with downstream 3D performance through a diagnostic investigation. (ii) We propose \textbf{\model}, a geometry-aware MLLM adaptation framework that integrates Camera Pose Injection, Multi-view Epipolar Attention, and 3D Geometric Distillation to strengthen spatial reasoning without relying on external 3D models at inference. (iii) Through \textbf{extensive experiments} on AV-oriented 3D scene understanding benchmarks, \model improves planning accuracy, spatial reasoning, and geometric understanding over strong MLLM baselines while maintaining efficient inference.
% \begin{itemize}[leftmargin=*,nosep]
% \item We identify depth estimation and cross-view correspondence as two key geometric bottlenecks for MLLM-based AV scene understanding, and validate their correlation with downstream 3D performance through a diagnostic investigation.
% \item We propose \textbf{\model}, a geometry-aware MLLM adaptation framework that integrates Camera Pose Injection, Multi-view Epipolar Attention, and 3D Geometric Distillation to strengthen spatial reasoning without relying on external 3D models at inference.
% \item Through extensive experiments on AV-oriented 3D scene understanding benchmarks, \textbf{\model} improves planning accuracy, spatial reasoning, and geometric understanding over strong MLLM baselines while maintaining efficient inference.
% \end{itemize}

% \nghicomment{[Nghi]: Clarify the exact novelty over OmniDrive, VGGDrive, and Spatial-MLLM. The introduction should explicitly state which baselines require test-time 3D modules or LiDAR/BEV-style intermediates, and why OmniSpace is simpler at deployment.}
% \input{sections/}
\section{Diagnostic Analysis: Geometry Awareness in Autonomous Driving MLLMs}
\label{sec:investigate}

% As aforementioned, Autonomous-driving MLLMs must reason beyond object recognition and scene description. Many critical safety tasks such as collision avoidance, lane-level decision, multi-view scene understanding, and trajectory planning, require metric depth, spatial layout, and cross-view correspondence. Thus, we investigate the following hypothesis: \emph{MLLMs with stronger depth estimation and cross-view correspondence should transfer better to downstream autonomous-driving tasks.}

\textbf{Why current MLLMs lack of geometry awareness. } As aforementioned, we hypothesize that the main bottleneck of current MLLMs lies in two specific abilities: depth estimation and cross-view correspondence. A likely cause is the pretraining paradigm itself. Most recent MLLMs inherit their visual representations from image--text pretraining under a CLIP-style paradigm~\cite{radford2021learning}. Such supervision is highly effective for learning semantic alignment between images and language, but limited explicit supervision for construct a geometric structure, and spatial layout. Consequently, the visual encoder may recognize objects correctly yet still fail to localize them in 3D space or to determine how observations from different cameras correspond to the same physical scene. To validate this observation and quantify its effect on driving performance, we conduct two complementary experiments described below.

\begin{figure*}[h]
\centering
\begin{minipage}[c]{0.6\linewidth}
  \centering
  \includegraphics[page=1, width=\linewidth]{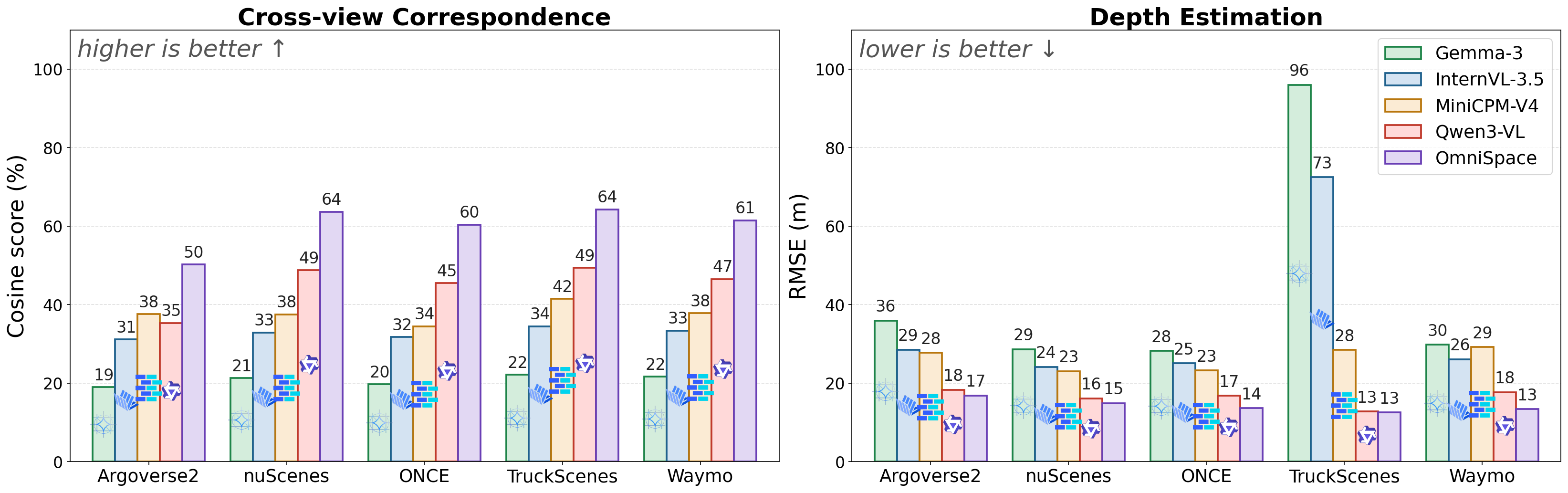}
\end{minipage}\hfill
\begin{minipage}[c]{0.38\linewidth}
  \centering
% shortcut
  \definecolor{qwencolor}{HTML}{FFD9D9}
  \newcommand{\hlq}{\cellcolor{qwencolor}}
  \aboverulesep=0pt
  \belowrulesep=0pt
  \setlength{\tabcolsep}{3pt}
  \renewcommand{\arraystretch}{1.2}
  \resizebox{\linewidth}{!}{%
  \begin{tabular}{cl c c c c}
  \toprule
  & \textbf{Metric}
  & \hlq \textbf{Qwen3-VL} & \textbf{InternVL3.5} & \textbf{Gemma-3} & \textbf{MiniCPM-V4} \\
  \midrule
  % ---------- nuInstruct ----------
  \multirow{5}{*}{\rotatebox[origin=c]{90}{\textbf{nuInstruct}}}
    & BLEU $\uparrow$     & \hlq 86.71 & 86.06 & 84.11 & 86.39 \\
    & Accuracy $\uparrow$ & \hlq 59.51 & 57.50 & 53.73 & 57.48 \\
    & MAP $\uparrow$      & \hlq 35.32 & 19.06 & 10.01 & 30.71 \\
    & MAE $\downarrow$    & \hlq 2.47  & 2.86  & 2.78  & 2.60 \\
    & \textit{Average} $\uparrow$ & \hlq 44.81 & 39.94 & 36.29 & 43.00 \\
  \midrule
  % ---------- nuScenes ----------
  \multirow{4}{*}{\rotatebox[origin=c]{90}{\textbf{nuScenes}}}
    & L2 $\downarrow$           & \hlq 0.36 & 0.37 & 0.37 & 0.37 \\
    & Collision $\downarrow$    & \hlq 0.36 & 0.35 & 0.39 & 0.35 \\
    & Intersection $\downarrow$ & \hlq 3.05 & 3.59 & 3.59 & 3.41 \\
    & \textit{Average} $\downarrow$ & \hlq 1.26 & 1.44 & 1.45 & 1.38 \\
  \midrule
  % ---------- Omnidrive ----------
  \multirow{4}{*}{\rotatebox[origin=c]{90}{\textbf{Omnidrive}}}
    & BLEU $\uparrow$  & \hlq 25.07 & 24.86 & 24.70 & 24.93 \\
    & ROUGE $\uparrow$ & \hlq 37.22 & 37.30 & 36.97 & 37.26 \\
    & CIDEr $\uparrow$ & \hlq 94.93 & 94.01 & 93.41 & 94.89 \\
    & \textit{Average} $\uparrow$ & \hlq 52.41 & 52.06 & 51.69 & 52.36 \\
  \bottomrule
  \end{tabular}%
  }
\end{minipage}

\caption{Correlation between cross-view correspondence, depth estimation, and downstream AV performance. \textbf{Left:} per-model scores on our diagnostic benchmark. \textbf{Right:} downstream results on nuInstruct, nuScenes, and OmniDrive after LoRA fine-tuning. }
\label{fig:motivation}
\end{figure*}

\noindent
\textbf{Experiment \#1 - Figure \ref{fig:motivation} (left):} We first conduct a diagnostic benchmark, called \textit{Geo-Bench}, sampled from five AV datasets: nuScenes~\cite{caesar2020nuscenes}, Argoverse~2~\cite{wilson2023argoverse}, Waymo~\cite{sun2020scalability}, ONCE~\cite{mao2021one}, and TruckScenes~\cite{fent2024man}. Geo-Bench targets two complementary geometric abilities, \textit{cross-view correspondence} and \textit{monocular depth understanding}, evaluated through distinct protocols to isolate whether model failures stem from inaccurate depth perception, weak multi-view association, or both. The QA pairs are designed specifically for depth understanding: each of the 250 multi-view questions asks the model to estimate the distance to a referenced object, and we report root mean squared error (RMSE) between the predicted and ground-truth distances. Cross-view correspondence, in contrast, is not tested through QA but through a representation-level probe: given a patch feature from one camera, we use LiDAR-derived depth and camera calibration to project its 3D location into another view and compute the cosine similarity between the two corresponding patch features, where higher similarity indicates that the model represents the same scene element consistently across views. Figure~\ref{fig:motivation} (left) reports the performance of four recent MLLMs~\cite{bai2025qwen3,team2024gemma,wang2025internvl3,yao2024minicpm} on Geo-Bench under both protocols. 
Details of the Geo-Bench construction pipeline are provided in Appendix~\ref{supp:data_construction}.

\noindent
\textbf{Experiment \#2 - Figure \ref{fig:motivation} (right):} We evaluate the same MLLMs on three AV benchmarks: nuInstruct, nuScenes, and OmniDrive~\cite{caesar2020nuscenes,ding2024holistic,wang2025omnidrive}. These benchmarks cover complementary aspects of driving intelligence, including risk perception, planning-related prediction, and driving-scene description. To adapt each MLLM to the target domain, we apply LoRA fine-tuning~\cite{hu2022lora} following the prompting protocol of~\cite{huang2025robotron}. We report the official metrics for each benchmark, including BLEU, accuracy, MAP, MAE, L2 error, collision rate, intersection rate, ROUGE, and CIDEr. \\
We first observe that generic MLLM capability does not always translate into strong AV-domain performance. For example, although InternVL3.5 is a strong general-purpose MLLM, MiniCPM-V4 outperforms it on several AV metrics. More importantly, we observe that \textit{geometry-aware models transfer better to driving tasks}. Models with stronger diagnostic geometry scores generally achieve better downstream AV performance. This supports our hypothesis that geometry awareness is not merely an auxiliary perception ability, but a core capability for autonomous-driving MLLMs. Motivated by this finding, we introduce a geometry-aware training strategy that strengthens monocular depth estimation and cross-view correspondence.

% Requires: \usepackage{booktabs, multirow, colortbl, xcolor, graphicx, array}
% Qwen column highlighted in light blue.

% Requires: \usepackage{booktabs, multirow, colortbl, xcolor, graphicx, array}
% Side-by-side: figure on the left, table on the right.
% - "Dataset" header removed.
% - No \cmidrule before the Average row inside each block.
% - Qwen column highlighted in light blue.

% Requires: \usepackage{booktabs, multirow, colortbl, xcolor, graphicx, array}
% Side-by-side: figure on the left, table on the right.
% - "Dataset" header removed.
% - No \cmidrule before the Average row inside each block.
% - Qwen column highlighted in light blue.

% Requires: \usepackage{booktabs, multirow, colortbl, xcolor, graphicx, array}
% Side-by-side: figure on the left, table on the right.
% - "Dataset" header removed.
% - No \cmidrule before the Average row inside each block.
% - Qwen column highlighted in light blue (cellcolor so it survives \resizebox).

% \nghicomment{[Nghi]: The diagnostic story is promising, but correlation across four MLLMs is not enough by itself. Add Pearson/Spearman correlations, confidence intervals, and a discussion of confounders such as model size, visual encoder, pretraining data, and LoRA setup.}
% \nghicomment{[Nghi]: Please spell out the diagnostic benchmark construction: object sampling, camera-pair selection, prompt templates, answer parsing, unit/range normalization, train-test separation, and whether any samples overlap with downstream training data.}
\section{\model}

% Motivated by the analysis in Section~\ref{sec:investigate}, we propose \textbf{\model}, a plug-in geometry-aware architecture that enhances 3D reasoning for autonomous-driving tasks across a broad range of MLLMs. \model is designed with three objectives: \textbf{(i)} inject camera-view spatial priors into visual tokens to improve metric depth awareness, \textbf{(ii)} encourage cross-view correspondence through epipolar-constrained interaction, and \textbf{(iii)} leverage 3D foundation models during training without relying on external geometry models at inference time.
% Figure~\ref{fig:main_figure} illustrates the overall architecture.
%In the following, we describe how each of these objectives is achieved.
Motivated by the analysis in Section~\ref{sec:investigate}, we proposed \textbf{\model}, a plug-in geometry-aware architecture to enhance 3D awareness for AV tasks across a broad range of MLLMs. \model is designed with three objectives: \textbf{(i)} inject spatial awareness from camera views into visual tokens to improve metric depth awareness, \textbf{(ii)} explicitly model encourage cross-view correspondence through epipolar-constrained interaction and \textbf{(iii)} leverage 3D foundation models during training without relying on external geometry models at inference time.

\paragraph{Preliminaries}
Given multi-view ($N$) images $\mathcal{I}_t=\{\mathbf{I}_t^1,\ldots,\mathbf{I}_t^N\}$ at timestamp $t$, a standard MLLM first encodes each image independently using a pretrained 2D visual encoder $\mathcal{E}_{2D}$. This produces visual feature maps $\mathbf{F}_t^n = \mathcal{E}_{2D}(\mathbf{I}_t^n) \in \mathbb{R}^{h \times w \times D}, \quad n=1,\ldots,N$. Generally, we call $\mathbf{F}^{2D}$. 
%The feature maps are flattened into visual tokens and projected into the language model token space through a lightweight connector. Together with instruction tokens from the text tokenizer, these visual tokens are consumed by the LLM backbone $f_\theta$ to generate the final response. Although effective for semantic understanding, this pipeline treats each camera view largely as an independent 2D image. The resulting visual tokens are weakly constrained by camera geometry and may not encode where a feature lies in 3D or how it corresponds to features in other views. \model addresses this limitation by augmenting the visual-token pathway with camera-pose injection, epipolar cross-view attention, and 3D geometric distillation.

% \hl{Combine 3.1 and 3.2 and devide into 2 steps}

\subsection{Camera Pose Injector}

\begin{figure}[t]
  \centering
  \includegraphics[page=1, width=\linewidth]{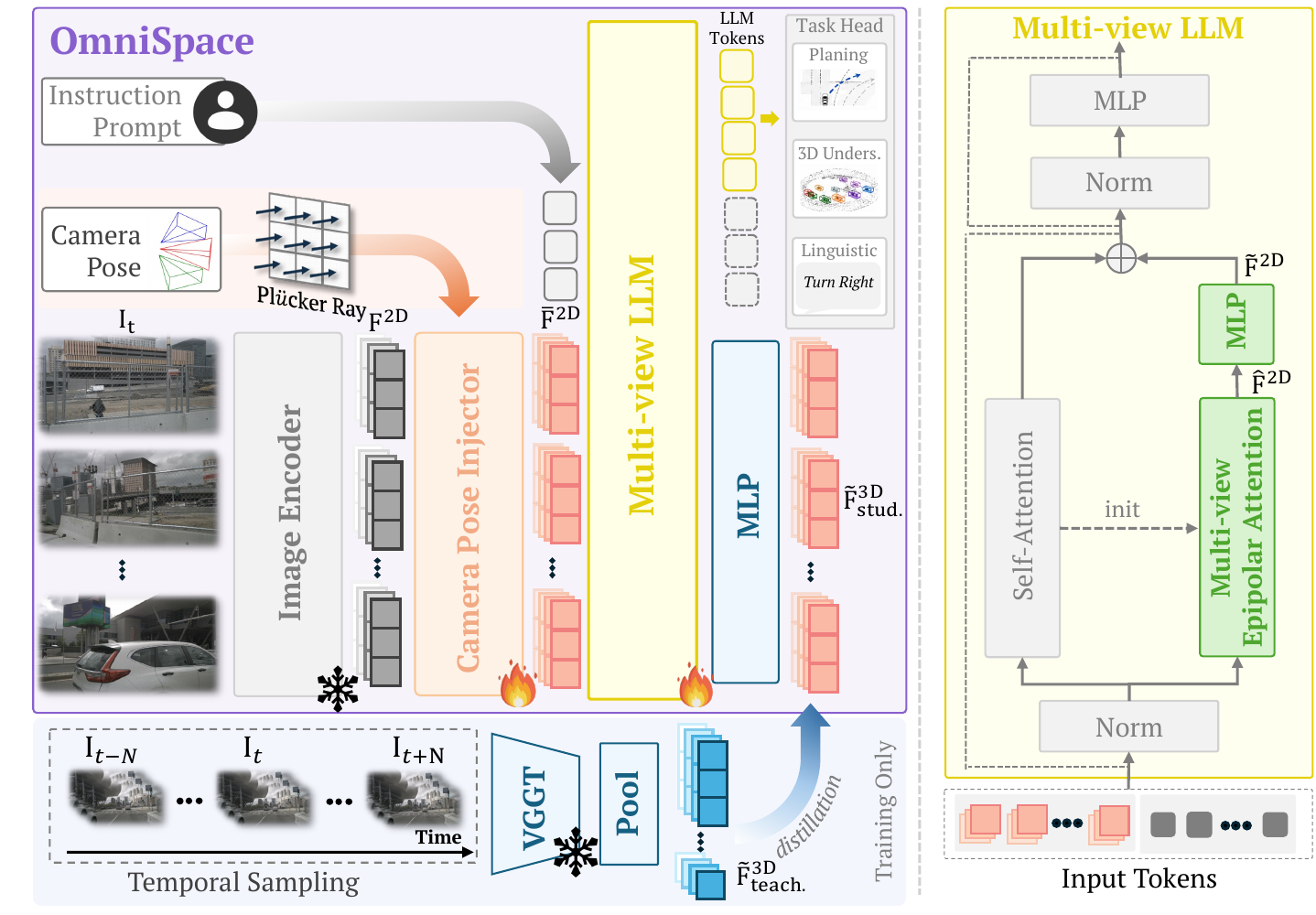}
    \caption{Overall architecture of the proposed \textbf{\model}. Given synchronized multi-view images, \model encodes 2D visual tokens with camera-pose-aware ray embeddings, refines them through epipolar-constrained cross-view attention, and distills 3D geometric priors during training. The 3D teacher is discarded at inference. }
    \label{fig:main_figure}
    \vspace{-10pt}
\end{figure}
We inject camera-aware spatial priors into each visual token. 
%The first component injects camera-aware spatial priors into each visual token. 
%To address these shortcomming, we introduce \textit{Camera Pose Injector}, a module designed to inject spatial awareness into MLLMs. 
Each point on the feature map $F^{2D}_{ij}$ corresponds to a camera ray. A ray is defined by a camera center $\mathbf{o}\in \mathbb{R}^3$ and a direction $\mathbf{d}$. As the na\"ive representation $(\mathbf{o}, \mathbf{d})$ is not ideal, because the same geometric ray can be written using different points along that ray: suppose we have one ray starts at camera center $\mathbf{o}$ and points in direction $\mathbf{d}$. We consider another point lies somewhere along the same ray $\mathbf{o'}=\mathbf{o} + t\mathbf{d}$, where $t$ is any scalar. If we represent the ray naively as ($\mathbf{o}$,$\mathbf{d}$), then ($\mathbf{o}, \mathbf{d}$) and ($\mathbf{o}+t\mathbf{d},\mathbf{d}$) look different, even though they describe the same geometric line. To avoid this problem, we are motivated by recent advances in Neural Light Fields~\cite{sitzmann2021light}, we adopt the Pl\"{u}cker ray embedding, ${P}_{ij} = (\mathbf{o}\times \mathbf{d}_{ij},\, \mathbf{d}_{ij}) \in \mathbb{R}^6$, where $\mathbf{d}_{ij}$ is the normalized world-space ray direction for token $(i, j)$. This parametrization elegantly maps $(\mathbf{o} \times \mathbf{d})$ and $(\mathbf{o}+t\mathbf{d} )\times \mathbf{d}$ to an identical embedding since any vector crossed with itself is zero:
\begin{equation}
    (\mathbf{o}+t\mathbf{d}_{ij})\times \mathbf{d}_{ij} = \mathbf{o} \times \mathbf{d}_{ij} + t\,(\mathbf{d}_{ij} \times \mathbf{d}_{ij}) = \mathbf{o} \times \mathbf{d}_{ij}.
\end{equation}
% Because any vector crossed with itself is zero, $\mathbf{d}_{ij} \times \mathbf{d}_{ij} = 0$.

%Concretely, consider a camera whose optical center is located at $o\in \mathbb{R}^3$ in the world coordinate system. For each point on the feature map $F^{2D}_{ij}$, we define the corresponding ray along a normalized direction $d\in \mathbb{R}^3$ as $r_{ij} = (o,d)$. This ray is then encoded as a positional signal, enabling the model to disambiguate views and to localize where a feature ``point'' resides in world coordinates. Unlike prior approaches~\cite{liu2022petr,wang2025omnidrive} that discretize $k$ 3D points along the ray $r_{ij}$, an approach that is both data-hungry and prone to depth ambiguity~\cite{liu2024ray}, our formulation avoids such sampling altogether. Moreover, consider two rays sharing the same direction but originating from distinct camera centers lying on the same line, $r^1_{ij} = (o,d)$ and $r^2_{ij} = (o+td,d)$ \hl{what is t?}. Although these rays are geometrically equivalent, na\"{i}ve embeddings assign them markedly different representations.

% Motivated by recent advances in Neural Light Fields~\cite{sitzmann2021light}, we instead adopt the Pl\"{u}cker ray embedding $\mathcal{P}_{ij} = (o\times d,\, d)$, where $\times$ denotes the cross product. This parametrization elegantly maps $r^1$ and $r^2$ to an identical embedding, since 
% \begin{equation}
%     (o+td)\times d = o \times d + t\,(d \times d) = o \times d.
% \end{equation}
The resulting Pl\"{u}cker ray map $\mathbf{P}\in \mathbb{R}^{(hw)\times 6}$ is passed through a lightweight projection $\phi_p$, forming pose-aware feature map: $\bar{F}_{ij}^{2D} =F_{ij}^{2D} + \phi_p(P_{ij})$, where $\phi_p$ is implemented as a $1\times1$ convolution. To avoid disrupting the pretrained visual representation at the beginning of fine-tuning, $\phi_p$ is zero-initialized. Therefore, \model initially behaves identically to the base MLLM and gradually learns to inject camera geometry as training proceeds.

%a $1\times 1$ convolutional layer that projects it to dimension $D$ and is then added to the multi-view feature maps $F_{2D}$. To preserve the integrity of the pre-trained backbone, the projection layer is initialized with zero weights and learned during fine-tuning, ensuring that spatial priors are introduced progressively without disrupting the original representations. Overall achieved our first objective.

% \subsection{Multi-view Epipolar Attention}
% \label{sec:Epipolar}
% \begin{wrapfigure}{r}{0.25\linewidth}
% \vspace{-15pt}
%   \centering
%   \includegraphics[width=\linewidth]{images/epipolar_attention.pdf}
% \vspace{-22pt}
%   \caption{Demostration of Multi-view Epipolar Attention.}
%   \label{fig:epipolar_attention}
% \vspace{-40pt}
% \end{wrapfigure}

%Even each token is injected with ray-level spatial context, it does not explicitly enforce cross-view correspondence. 
\subsection{Multi-view Epipolar Attention}
\label{sec:Epipolar}
To encourage view-consistent representation learning, we propose to inject camera-aware geometry into cross-view attention so that a token in one view can only attend to geometrically plausible tokens in another view, rather than attending to every token everywhere. Let $\bar{\mathbf{F}}^{2D}_{i}$ be the pose-aware feature map from view $i$, $s$ is one source token location in that feature map, we compute its epipolar lines $\{l_j\}_{j \neq i}$ on all other pose-aware feature maps $\bar{\mathbf{F}}^{2D}_j\}_{j \neq i}$. When computing the attention map across views, we ignore points that do not lie on these epipolar lines, so that the source point $s$ only attends to features along its corresponding camera ray in other views, in addition to all points within its own view:
\begin{equation}
\hat{\mathbf{F}}^{2D}_{i,s} = \mathrm{SoftMax}\left(\frac{Q(\bar{\mathbf{F}}^{2D}_{i,s}) \cdot K([\bar{\mathbf{F}}^{2D}_{i} \mid \bar{\mathbf{F}}^{2D}_{j,l_j}])^\top}{\sqrt{d}}\right) \cdot V([\bar{\mathbf{F}}^{2D}_{i} \mid \bar{\mathbf{F}}^{2D}_{j,l_j}]).
\end{equation}
where $\bar{\mathbf{F}}^{2D}_{i,s}$ is the visual token at position $s$ in view $i$ and $Q\bar{\mathbf{F}}^{2D}_{i,s}$ is the query vector of the source token. $[\mathbf{F}^{2D}_{i} \mid \mathbf{F}^{2D}_{j,l_j}]$ is set of tokens that the source token is allowed to attend to whereas $\mathbf{F}^{2D}_{i}$ contains all tokens from the same view and $\bar{\mathbf{F}}^{2D}_{j,l_j}$ is only the tokens in another view $j$ that lie on the epipolar line $l_j$. The operator $\mid$ denotes concatenation along the token dimension. $Q$, $K$, and $V$ are learnable linear projections that map visual tokens into query, key, and value embeddings, respectively. In this way, each token preserves unrestricted self-attention within its own view while aggregating cross-view evidence only from epipolar-consistent regions. In practice, we dilate each epipolar line with a $3 \times 3$ filter to include neighboring target tokens and improve robustness to feature-map discretization and calibration noise. Finally, we inject the epipolar-attended feature through a zero-initialized projection layer: $\tilde{\mathbf{F}}^{2D}_{i,s} = \phi_{f}
\hat{\mathbf{F}}^{2D}_{i,s},$
where $\phi_{f}$ is initialized to zero and learned during fine-tuning. This residual design allows the model to gradually acquire geometry-aware cross-view reasoning while preserving the behavior of the original pretrained model at initialization.

\subsection{3D Geometric Distillation}

\begin{wrapfigure}{tr}{0.5\linewidth}
    \vspace{-10pt}
  \centering
  \includegraphics[ width=\linewidth]{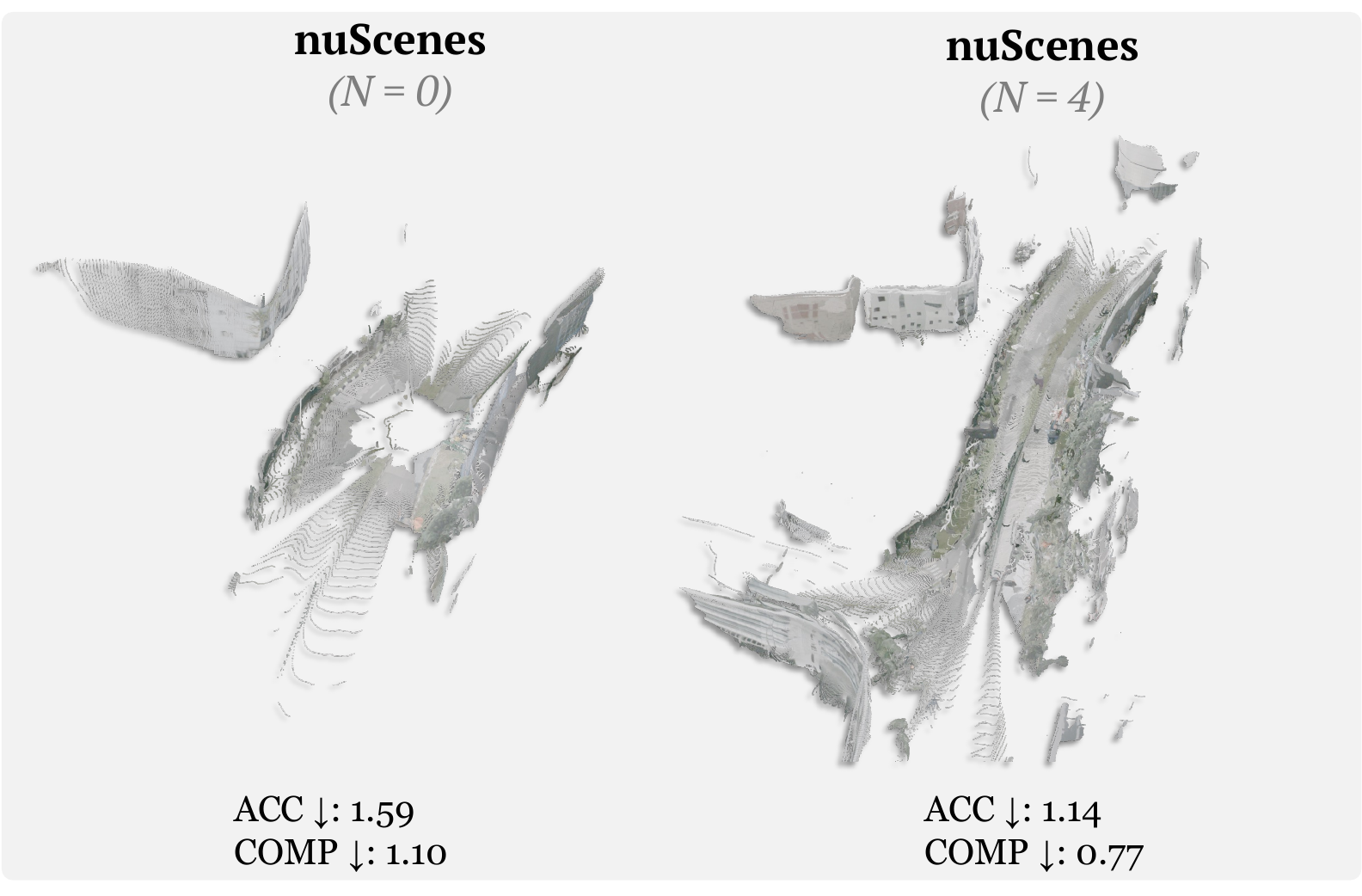}
  % \caption{Qualitative and quatitative comparision of VGGT~\cite{wang2025vggt} on nuScenes~\cite{caesar2020nuscenes} and ETH3D~\cite{schops2017multi}.}
  \caption{Qualitative and quantitative results of VGGT~\cite{wang2025vggt} on nuScenes~\cite{caesar2020nuscenes} validation sets under current-frame and temporally aggregated inputs.}
  \label{fig:vggt_issue}

    \vspace{-12pt}
\end{wrapfigure}

While camera pose injection and multi-view epipolar attention introduce explicit geometric structure into the visual-token pathway, they do not by themselves specify what a 3D-aware representation should encode. A straightforward solution is to use LiDAR or an external 3D foundation model at inference time, but this increases deployment cost and may introduce cascading errors when the external model fails. Instead, we leverage a feed-forward 3D foundation model~\cite{wang2025vggt}, denoted by $\mathcal{E}_{3D}$, only during training and distill its geometric knowledge into \model.

Naively following Spatial-MLLM~\cite{wu2025spatial} by feeding only synchronized multi-view images at timestamp $t$, $\mathcal{I}_t$, into $\mathcal{E}_{3D}$ is suboptimal for AV scenes. Existing feed-forward 3D models are typically trained on indoor, object-centric, or reconstruction-oriented datasets~\cite{jensen2014large,schops2017multi,zhou2018stereo,reizenstein2021common}, where views have substantial overlap and scenes are relatively static. In contrast, AV scenes contain limited inter-view overlap, large outdoor depth ranges, moving objects, and frequent occlusions. As shown in Figure~\ref{fig:vggt_issue}, directly applying $\mathcal{E}_{3D}$ to current-frame AV inputs can produce noisy and incomplete geometric supervision.

%To obtain more reliable 3D targets, 
To ensure reliability, 
we introduce \emph{temporal geometric aggregation}. At timestep $t$, we sample a symmetric temporal window of $2K+1$ synchronized multi-view frames: $\mathcal{C}_t = \{\mathcal{I}_{t-K}, \ldots, \mathcal{I}_{t}, \ldots, \mathcal{I}_{t+K}\}.$
% \begin{equation}
%     \mathcal{C}_t = \{\mathcal{I}_{t-K}, \ldots, \mathcal{I}_{t}, \ldots, \mathcal{I}_{t+K}\}.
% \end{equation}
We pass the full clip into the 3D teacher $\mathcal{E}_{3D}$ and obtain $\mathcal{E}_{3D}(\mathcal{C}_t) = \{\mathbf{F}^{3D}_{t-K}, \ldots, \mathbf{F}^{3D}_{t}, \ldots, \mathbf{F}^{3D}_{t+K}\}$, where $\mathbf{F}^{3D}_{\tau}\in\mathbb{R}^{HW\times \hat{D}}$ denotes the teacher feature map at timestamp $\tau$. We aggregate the teacher features along the temporal dimension to form a stable teacher:
\begin{equation}
    \tilde{\mathbf{F}}^{3D}_{teacher} = \frac{1}{2K+1} \sum_{\tau=t-K}^{t+K} \mathbf{F}^{3D}_{\tau} \in \mathbb{R}^{HW\times \hat{D}}.
\end{equation}
This temporal aggregation increases effective scene overlap across frames and reduces single-frame reconstruction artifacts caused by sparse camera overlap or transient occlusions.

% \nghicomment{[Nghi]: Temporal averaging assumes teacher features are aligned across time. Please explain whether ego-motion compensation or world-frame alignment is used, and how moving objects are handled.}

Given the instruction prompt $T$ and geometry-enhanced visual tokens $\tilde{\mathbf{F}}^{2D}$ from Section~\ref{sec:Epipolar}, the MLLM backbone produces visual representations $\rttensortwo{\mathbf{F}}^{2D}$. We project these student representations into the teacher feature dimension using a lightweight MLP: $\tilde{\mathbf{F}}^{3D}_{stud.} = \phi_d(\rttensortwo{\mathbf{F}}^{2D}) \in \mathbb{R}^{hw\times \hat{D}}.$
% \begin{equation}
%     \mathbf{F}^{2D}_{stud.} = \phi_d(\rttensortwo{\mathbf{F}}^{2D}) \in \mathbb{R}^{hw\times \hat{D}}.
% \end{equation}

Since the teacher and student feature maps may have different spatial resolutions, we apply adaptive pooling to align the teacher feature map to the student resolution. The distillation objective is:
\begin{equation}
    \mathcal{L}_{3D} = \left\| \tilde{\mathbf{F}}^{3D}_{stud.}  - \mathrm{Pool}(\tilde{\mathbf{F}}^{3D}_{teach.}) \right\|_2^2.
    \label{eq:distill}
\end{equation}
This objective encourages the MLLM visual tokens to encode 3D-aware structure learned from the teacher $\mathcal{E}_{3D}$, while preserving the original language-modeling objective for downstream instruction driving reasoning. Because $\mathcal{E}_{3D}$ is used only to generate training targets and is discarded at inference, \model inherits 3D geometric priors without test-time dependence on external 3D models.

%\subsection{Training Objectives}
The overall training objective combines the standard downstream cross-entropy loss with the 3D geometric distillation loss defined in Eq.~\ref{eq:distill}.
% \nghicomment{[Nghi]: Add the full training objective here, including the language loss, the 3D distillation weight, whether the teacher is frozen, and which student layer/tokens are supervised.}

\section{Experiments}

\begin{table*}[t]
    \centering
    \caption{\textbf{Open-loop planning results on nuScenes}. The best is \textbf{bold} and the runner up is \underline{underline}.}
    \label{tab:open_loop}
    \resizebox{\textwidth}{!}{
        \begin{tabular}{c|l|c|c|c|cccc|cccc|cccc}        \toprule
        \multicolumn{1}{c|}{} & \multirow{2}{*}{Method} & \multirow{2}{*}{LLM} & \multirow{2}{*}{Params} & \multirow{2}{*}{FPS $\uparrow$} & \multicolumn{4}{c|}{L2 $\downarrow$} & \multicolumn{4}{c|}{Collision (\%) $\downarrow$} & \multicolumn{4}{c}{Intersection (\%) $\downarrow$} \\
        \cmidrule(lr){6-9} \cmidrule(lr){10-13} \cmidrule(lr){14-17}        
        \multicolumn{1}{c|}{} & & & & & 1s & 2s & 3s & Avg. & 1s & 2s & 3s & Avg. & 1s & 2s & 3s & Avg.  \\
        \midrule
        \multirow{4}{*}{\rotatebox[origin=c]{90}{Traditional}}
        & UniAD \cite{hu2023planning}       & -- & 125M & 1.8 & 0.20 & 0.42 & 0.75 & 0.46 & 0.02 & 0.25 & 0.84 & 0.37 & \textbf{0.20} & 1.33 & 3.24 & 1.59 \\
        & VAD-Base \cite{jiang2023vad}     & -- & 58M & 4.5 & 0.17 & 0.34 & 0.60 & 0.37 & 0.04 & 0.27 & 0.67 & 0.33 & \underline{0.21} & 2.13 & 5.06 & 2.47 \\
        & BEV-Planner \cite{li2024ego}      & -- & -- & -- & 0.16 & 0.32 & 0.57 & 0.35 & \textbf{0.00} & 0.29 & 0.73 & 0.34 & 0.35 & 2.62 & 6.51 & 3.16 \\
        & ST-P3~\cite{hu2022st}             & -- & 11M & -- & 1.59 & 2.64 & 3.73 & 2.65 & 0.69 & 3.62 & 8.39 & 4.23 & 2.53 & 8.17 & 14.40 & 8.37 \\

        \midrule
        \multirow{15}{*}{\rotatebox[origin=c]{90}{MLLMs-based}}
        % ===================== Group 1: 2D MLLM (vision-language only) =====================
        & \multicolumn{16}{l}{\cellcolor{gray!15}\textit{2D MLLM }} \\
        & Baseline                          & Qwen2.5-VL 7B  & 7.75B & 0.42 & 0.15 & 0.33 & 0.62 & 0.37 & \textbf{0.00} & 0.18 & 0.76 & 0.31 & 0.94 & 3.09 & 6.76 & 3.59 \\
        & Baseline                          & InternVL3.5 4B & 4.73B & 0.47 & 0.15 & 0.33 & 0.62 & 0.37 & \textbf{0.00} & 0.18 & 0.88 & 0.35 & 0.78 & 3.03 & 6.95 & 3.59 \\
        & Baseline                          & Qwen3-VL 4B    & 4.4B  & 0.44 & 0.15 & 0.32 & 0.61 & 0.36 & \textbf{0.00} & 0.14 & 0.96 & 0.36 & 0.57 & 2.54 & 6.04 & 3.05 \\
        & EMMA~\cite{hwang2024emma}         & --             & --    & --   & 0.14 & 0.29 & 0.54 & 0.32 & -    & -    & -    & -    & -    & -    & -    & -    \\
        & DriveVLM~\cite{tian2025drivevlm}  & Qwen-VL        & 9.6B  & --   & 0.18 & 0.34 & 0.68 & 0.40 & 0.10 & 0.22 & \underline{0.45} & 0.27 & -    & -    & -    & -    \\
        % ===================== Group 2: 2D MLLM + test-time 3D model =====================
        & \multicolumn{16}{l}{\cellcolor{gray!15}\textit{2D MLLM + test-time 3D model}} \\
        & ORION~\cite{fu2025orion}          & Vicuna-7B-v1.5 & 7.55B & --   & 0.17 & 0.31 & 0.55 & 0.34 & 0.05 & 0.25 & 0.80 & 0.37 & -    & -    & -    & -    \\
        & OmniDrive \cite{wang2025omnidrive}& LLaVA-7B       & 7.2B  & 0.27 & 0.14 & 0.29 & 0.55 & 0.33 & \textbf{0.00} & \underline{0.13} & 0.78 & 0.30 & 0.56 & 2.48 & 5.96 & 3.00 \\
        & VGGDrive \cite{wang2026vggdrive}  & Qwen2.5-VL 7B  & 8.5B  & 0.03 & 0.14 & 0.28 & 0.51 & 0.31 & 0.02 & \textbf{0.10} & 0.55 & 0.22 & 0.63 & 2.27 & 4.02 & 2.31 \\
        & SpaceDrive~\cite{li2025spacedrive}& Qwen2.5-VL 7B  & 8.7B  & 0.25 & 0.15 & 0.29 & 0.51 & 0.32 & 0.04 & 0.18 & 0.49 & 0.23 & 0.22 & \textbf{0.80} & 2.79 & \underline{1.27} \\
        % ===================== Group 3: 2D MLLM + train-time 3D model (Ours) =====================
        & \multicolumn{16}{l}{\cellcolor{gray!15}\textit{2D MLLM + train-time 3D model}} \\
        & \cellcolor{qwenblue} \textbf{\ourmodule} & \cellcolor{qwenblue} Qwen2.5-VL 7B  & \cellcolor{qwenblue} 7.79B & \cellcolor{qwenblue} 0.40 & \cellcolor{qwenblue} \textbf{0.12} & \cellcolor{qwenblue} \textbf{0.25} & \cellcolor{qwenblue} \underline{0.47} & \cellcolor{qwenblue} \textbf{0.28} & \cellcolor{qwenblue} \textbf{0.00} & \cellcolor{qwenblue} \textbf{0.10} & \cellcolor{qwenblue} 0.57 & \cellcolor{qwenblue} 0.22 & \cellcolor{qwenblue} 0.32 & \cellcolor{qwenblue} 1.22 & \cellcolor{qwenblue} \textbf{2.22} & \cellcolor{qwenblue} \textbf{1.25} \\
        & \cellcolor{qwenblue} \textbf{\ourmodule} & \cellcolor{qwenblue} InternVL3.5 4B & \cellcolor{qwenblue} 4.77B & \cellcolor{qwenblue} 0.46 & \cellcolor{qwenblue} \underline{0.13} & \cellcolor{qwenblue} \underline{0.26} & \cellcolor{qwenblue} 0.50 & \cellcolor{qwenblue} \underline{0.30} & \cellcolor{qwenblue} \underline{0.01} & \cellcolor{qwenblue} \textbf{0.10} & \cellcolor{qwenblue} 0.50 & \cellcolor{qwenblue} \underline{0.20} & \cellcolor{qwenblue} 0.31 & \cellcolor{qwenblue} 1.88 & \cellcolor{qwenblue} \underline{2.32} & \cellcolor{qwenblue} 1.50 \\
        & \cellcolor{qwenblue} \textbf{\ourmodule} & \cellcolor{qwenblue} Qwen3-VL 4B    & \cellcolor{qwenblue} 4.44B & \cellcolor{qwenblue} 0.42 & \cellcolor{qwenblue} \textbf{0.12} & \cellcolor{qwenblue} \underline{0.26} & \cellcolor{qwenblue} \textbf{0.45} & \cellcolor{qwenblue} \textbf{0.28} & \cellcolor{qwenblue} 0.02 & \cellcolor{qwenblue} 0.14 & \cellcolor{qwenblue} \textbf{0.42} & \cellcolor{qwenblue} \textbf{0.19} & \cellcolor{qwenblue} 0.25 & \cellcolor{qwenblue} \underline{1.20} & \cellcolor{qwenblue} 2.36 & \cellcolor{qwenblue} \underline{1.27} \\
        
        \bottomrule
    \end{tabular}
    }
\end{table*}
% \nghicomment{[Nghi]: Use controlled same-backbone comparisons and report variance or statistical significance for small planning gains. Mixing LLM sizes, training data, sensors, and inference pipelines makes SOTA comparisons hard to interpret.}

% \nghicomment{[Nghi]: BLEU/ROUGE/CIDEr mainly measure language overlap, not geometry. Add geometry-sensitive metrics or qualitative failure cases to support spatial-reasoning claims on OmniDrive.}

% \nghicomment{[Nghi]: Add the missing ablations reviewers will expect: naive ray vs Plucker ray, single-frame teacher vs temporal teacher, different temporal windows, alternate teachers, vanilla vs epipolar attention under identical training, and calibration-noise robustness.}

% \nghicomment{[Nghi]: Sharpen this comparison by separating methods that use external 3D modules at inference, methods that rely on LiDAR/BEV/query tokens, and camera-only methods. This will make the deployment advantage of OmniSpace clearer.}

% \nghicomment{[Nghi]: Clarify the novelty relative to Spatial-MLLM, VGGDrive, LLaVA-3D, and 3DRS. Reviewers will ask whether the contribution is mainly ray embeddings plus teacher distillation, so the related work should explicitly distinguish the technical and deployment differences.}

\subsection{Experiment Setup}
\noindent \textbf{Datasets and Metrics.} 
We evaluate \model on five autonomous-driving benchmarks spanning trajectory planning (nuScenes~\cite{caesar2020nuscenes} and Bench2Drive~\cite{jia2024bench2drive}), risk detection (NuInstruct~\cite{ding2024holistic}), language understanding (OmniDrive~\cite{wang2025omnidrive}), and generalization (DriveBench~\cite{xie2025vlms}). For fair comparison, each model is trained and evaluated following the official protocol and metrics of each benchmark. 

% Dataset statistics and metric details are provided in Appendix~\ref{supp:dataset_and_metric}.

%To comprehensively evaluate \model's effectiveness in enhancing MLLMs with geometric awareness, we conduct experiments on five mainstream autonomous driving benchmarks. These include two trajectory planning benchmarks, nuScenes~\cite{caesar2020nuscenes} and Bench2Drive~\cite{jia2024bench2drive}, as well as three benchmarks covering risk detection, language, and generalization: NuInstruct~\cite{ding2024holistic}, OmniDrive~\cite{wang2025omnidrive}, and DriveBench~\cite{xie2025vlms}. To ensure a fair comparison and fully reflect the advantages of \model, we train and evaluate it on each dataset using its corresponding official metrics. Detailed statistics of the training sets and evaluation metrics are provided in Appendix~\ref{supp:dataset_and_metric}. 

% \subsection{Implementation Details}
\noindent \textbf{Baselines and Implementation Details.}
We evaluate \model with several MLLM backbones, including Qwen3-VL-4B~\cite{bai2025qwen3}, Qwen2.5-VL-7B~\cite{bai2025qwen2}, and InternVL3.5-4B~\cite{wang2025internvl3}, using Qwen3-VL-4B as the default backbone. All backbones are fine-tuned with LoRA~\cite{hu2022lora} using rank 16, while the original vision encoder and vision-language projector frozen. We use VGGT~\cite{wang2025vggt} as the 3D teacher $\mathcal{E}_{3D}$ and set the temporal sampling window parameter to $N=4$. Models are trained for 2 epochs with batch size 8, learning rate $1\times10^{-4}$, and cosine annealing. For fair comparison, we follow~\cite{wang2026vggdrive,wang2025omnidrive,li2025spacedrive} and incorporate ego-vehicle states into the instruction prompt for planning tasks. All experiments are conducted on 8 NVIDIA A100 GPUs 40GB. %Our experiments leverage several MLLMs, including Qwen3-VL 4B~\cite{bai2025qwen3}, Qwen2.5-VL 7B~\cite{bai2025qwen2}, and InternVL3.5 4B~\cite{wang2025internvl3}, with Qwen3-VL 4B serving as our default model. We finetune the core MLLMs using LoRA~\cite{hu2022lora} with a rank of 16, while keeping the original vision encoder and vision-language projector frozen. VGGT~\cite{wang2025vggt} is adopted as our teacher model for geometric distillation without further finetuning, and the number of sampling parameters $N$ is set to 4. The model is trained for 2 epochs with a batch size of 8. The learning rate is set to $1\times10^{-4}$, and cosine annealing is employed to ensure stable training. For the planning task, we follow~\cite{wang2026vggdrive,wang2025omnidrive,li2025spacedrive} incorporate the ego-vehicle state (e.g., velocity and acceleration) and command information (e.g., \textit{Go Straight}, \textit{Turn Left}, \textit{Turn Right}) into the instruction prompt. All experiments are conducted on 8 $\times$ NVIDIA A100 GPUs. 

\subsection{State-of-the-art Comparison}

\noindent \textbf{Evaluation on Open-loop Planning on nuScenes.} Table~\ref{tab:open_loop} shows that \ourmodule consistently improves over its corresponding fine-tuned baselines across Qwen2.5-VL, InternVL3.5, and Qwen3-VL backbones, indicating that the gains come from the proposed spatial modeling rather than a specific backbone. Compared with MLLM-based methods that rely on test-time 3D modules, \ourmodule achieves competitive or better planning accuracy while maintaining substantially higher throughput. These results suggest that training-time geometric supervision and explicit camera-pose modeling provide an effective accuracy--efficiency trade-off for open-loop planning.

%\noindent \textbf{Evaluation on Open-loop Planning on nuScenes.} As shown in Table~\ref{tab:open_loop}, \ourmodule consistently surpasses existing MLLM-based methods across all reported metrics. The lowest average L2 error (0.28) indicates that coordinate-level regression, guided by enhanced spatial priors, enables closer adherence to expert driving trajectories, while the markedly reduced Collision (0.19\%) and Intersection (1.27\%) rates further demonstrate that OmniSpace excels not only at fitting ground-truth trajectories but also at comprehensively enhancing driving safety through superior spatial understanding and reasoning. To ensure a fair comparison under matched settings, we further instantiate OmniSpace with multiple MLLM backbones (Qwen2.5-VL 7B, InternVL3.5 4B, and Qwen3-VL 4B), and observe consistent improvements over their respective fine-tuning baselines, indicating that the gains stem from our spatial modeling design rather than from a particular backbone choice. Beyond accuracy, OmniSpace also exhibits a favorable efficiency trade-off: whereas competing approaches such as OmniDrive (0.27 FPS), VGGDrive (0.03 FPS), and SpaceDrive (0.25 FPS) rely on auxiliary 3D modules or external geometric backbones that incur substantial inference overhead, OmniSpace sustains a markedly higher throughput (0.40--0.46 FPS across LLM backbones), approaching the latency of unmodified fine-tuning baselines while delivering superior planning performance.

\begin{table*}[t]
    \caption{
    \textbf{Closed-loop planning results on Bench2Drive.}
    The best is \textbf{bold} and the runner up is \underline{underline}. Methods marked with $^{\dagger}$ use additional training data beyond the standard benchmark.}
    \label{tab:closed_loop_benchmark}
    \vspace{-5pt}
    \centering
    % ============== LEFT: Traditional methods ==============
    \begin{minipage}[t]{0.45\linewidth}
    \vspace{0pt}
    \centering
    \resizebox{\linewidth}{!}{
    \begin{tabular}{l|c|cc}
    \toprule
    Method & Params & DS $\uparrow$ & SR(\%) $\uparrow$ \\
    \midrule
    \rowcolor{gray!15}
    \multicolumn{4}{l}{\textit{Traditional}} \\
    AD-MLP~\cite{zhai2023rethinking}                  & 285K & 18.05 & 0.00  \\ 
    UniAD-Base~\cite{hu2023planning}                  & 125M & 45.81 & 16.36 \\ 
    VAD-Base~\cite{jiang2023vad}                      & 58M  & 42.35 & 15.00 \\ 
    MomAD~\cite{song2025don}                          & 87M  & 44.54 & 16.71 \\ 
    GenAD~\cite{zheng2024genad}                       & 113M & 44.81 & 15.90 \\ 
    SparseDrive~\cite{sun2025sparsedrive}             & 86M  & 47.38 & 17.72 \\ 
    UAD~\cite{guo2025end}                             & --   & 49.22 & 20.45 \\ 
    WoTE~\cite{li2025end}                             & 64M  & 61.71 & 31.36 \\ 
    ThinkTwice~\cite{jia2023think}                    & 129M & 62.44 & 37.17 \\ 
    DriveTransformer-L~\cite{jia2025drivetransformer} & 520M & 63.46 & 38.60 \\ 
    DriveAdapter~\cite{jia2023driveadapter}           & 161M & 64.22 & 42.08 \\ 
    HiP-AD~\cite{tang2025hipad}                       & 98M  & 86.77 & 69.09 \\ 
    \bottomrule
    \end{tabular}}
    \end{minipage}%
    \hspace{0.01\linewidth}%
    % ============== RIGHT: MLLM-based methods ==============
    \begin{minipage}[t]{0.465\linewidth}
    \vspace{0pt}
    \centering
    \resizebox{\linewidth}{!}{
    \begin{tabular}{l|c|c|cc}
    \toprule
    Method & LLM & Params & DS $\uparrow$ & SR(\%) $\uparrow$ \\
    \midrule
    % --- Group 1: 2D MLLM ---
    \rowcolor{gray!15}
    \multicolumn{5}{l}{\textit{2D MLLM}} \\
    ReAL-AD~\cite{lu2025real}     & MiniCPMLlama3-2.5V  & --    & 41.17 & 11.36 \\ 
    Dual-AEB~\cite{zhang2025dual} & Qwen-0.5B           & --    & 45.23 & 10.00 \\ 
    X-Driver~\cite{liu2025x}      & LLaVA               & --    & 51.70 & 18.10 \\ 
    GEMINUS~\cite{wan2025geminus} & --                  & --    & 65.39 & 37.73 \\ 
    VDRive~\cite{guo2025vdrive}   & InternVL3-8B        & --    & 66.25 & 50.51 \\ 
    StuckSolver~\cite{bao2025large}                & GPT-4o              & --    & 70.89 & 50.01 \\ 
    DriveMoE~\cite{yang2025drivemoe}               & Paligemma-3B        & --    & 74.22 & 48.64 \\ 
    ETA~\cite{hamdan2025eta}      & LLaVA 1.6 7B Vicuna & --    & 74.33 & 48.33 \\ 
    VLR-Drive~\cite{kong2025vlr}  & LLaVA-NeXT Video 7B & --    & 75.01 & 50.00 \\ 
    \textcolor{gray!70}{SimLingo~\cite{renz2025simlingo}$^{\dagger}$} & \textcolor{gray!70}{InternVL2-1B} & \textcolor{gray!70}{1.36B} & \textcolor{gray!70}{85.07} & \textcolor{gray!70}{67.27} \\
    % --- Group 2: 2D MLLM + test-time 3D model ---
    \rowcolor{gray!15}
    \multicolumn{5}{l}{\textit{2D MLLM + test-time 3D model}} \\
    ORION~\cite{fu2025orion}              & Vicuna-7B-v1.5 & 7.55B & \underline{77.74} & 54.62 \\ 
    SpaceDrive~\cite{li2025spacedrive}    & Qwen2.5-VL 7B  & 8.7B  & 78.02 & 55.11 \\ 
    % --- Group 3: 2D MLLM + train-time 3D model (Ours) ---
    \rowcolor{gray!15}
    \multicolumn{5}{l}{\textit{2D MLLM + train-time 3D model (Ours)}} \\
    \rowcolor{qwenblue} \textbf{\model} & Qwen2.5-VL 7B & 7.79B & 79.65 & \underline{60.15} \\
    \rowcolor{qwenblue} \textbf{\model} & Qwen3-VL 4B   & 4.44B & \textbf{81.40} & \textbf{65.52} \\
    \bottomrule
    \end{tabular}}
    \end{minipage}
    \vspace{-10pt}
\end{table*}

\noindent \textbf{Evaluation on Closed-loop Planning on Bench2Drive.} In addition to the open-loop setting, we further conduct closed-loop evaluation to establish a more comprehensive and reliable assessment of planning performance. As shown in Table~\ref{tab:closed_loop_benchmark}, by introducing explicit geometric constraints, \model achieves a Driving Score of 79.65 and a Success Rate of 60.15\% with the Qwen2.5-VL backbone, surpassing SpaceDrive~\cite{li2025spacedrive} which relies on an external depth estimation model. A consistent trend is observed with the Qwen3-VL 4B variant, which ranks as the best-performing MLLM-based method on the benchmark, despite using a substantially smaller backbone than most competing approaches. Notably, although SimLingo~\cite{renz2025simlingo} reports a higher score, it benefits from extensive data augmentation through Action Dreaming, whereas \model attains comparable performance under the standard benchmark setting. These results collectively demonstrate the generalizability and effectiveness of our spatial-aware design across different backbones and evaluation regimes.

% \begin{table*}[htbp]
% 	\centering
% 	\tabcolsep=0.08cm
% 	\renewcommand{\arraystretch}{0.98}
% 	\caption{The performance comparison on the OmniDrive dataset \cite{wang2025omnidrive} focuses on caption-related tasks in which base VLMs excel.} \label{tab-4}
% 	\scalebox{0.78}{
%     \begin{tabular}{lccc cc c>{\columncolor{qwenblue}[\tabcolsep]}c}
%         \toprule
%         \multirow{2}{*}{Metrics} & \multicolumn{3}{c}{Zero-shot} & \multicolumn{4}{c}{Fine-tuned} \\
%         \cmidrule(lr){2-4} \cmidrule(lr){5-8}
%         & GPT-4o & LLAVA-OV & RoboTron & OmniDrive & HERMES & VGGDrive & \textbf{\model} \\
%         \midrule
%         BLEU↑   & 10.91 & 16.14 & 20.30 & 38.00 & -     & 37.58 & 37.88 \\
%         CIDEr↑  & 24.42 & 28.41 & 34.33 & 68.60 & 74.10 & 86.57 & 96.70 \\
%         ROUGE↑  & 22.34 & 22.14 & 23.67 & 32.60 & 32.70 & 34.40 & 40.44 \\
%         \textbf{Average↑} & 19.22 & 22.23 & 26.10 & 46.40 & - & 52.85 & 58.34 \\
%         \bottomrule
% 		\end{tabular}%
% 	}
% \end{table*}
\begin{wrapfigure}{r}{0.45\linewidth}
    \vspace{-2mm}
    \centering
    \includegraphics[width=\linewidth]{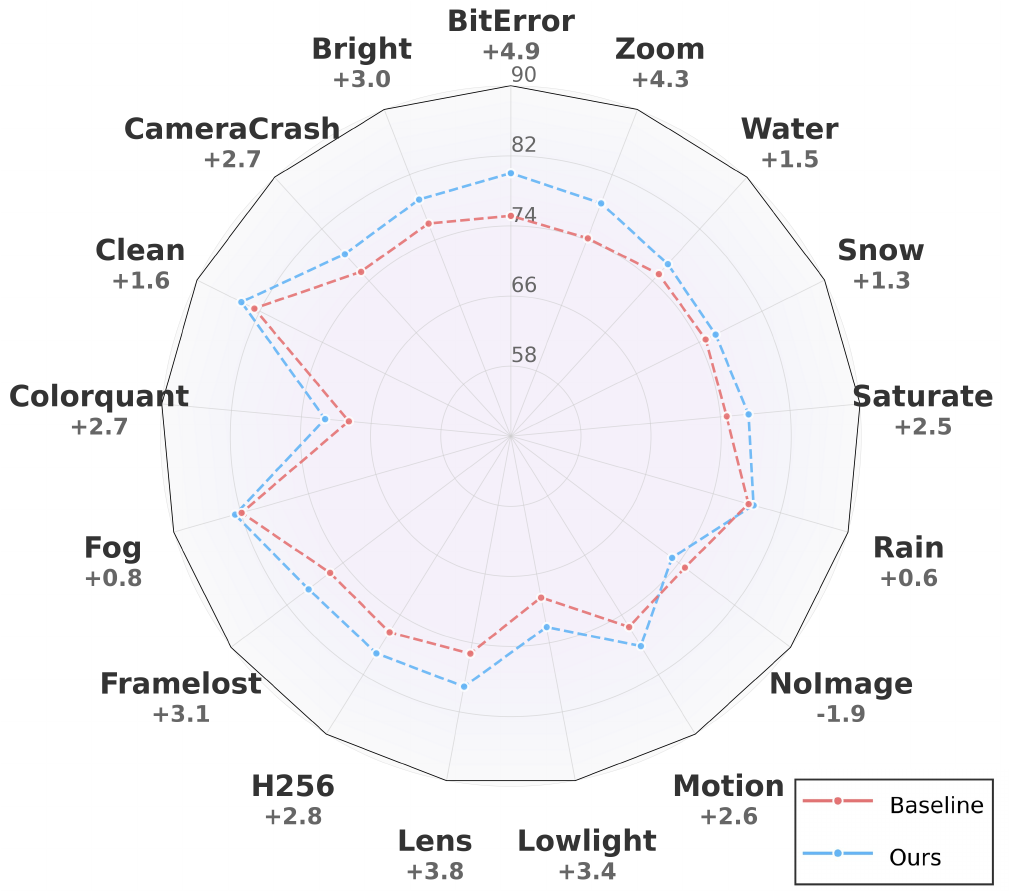}
    \vspace{-6mm}
    \caption{Generalization comparison on DriveBench~\cite{xie2025vlms}, evaluated using GPT-Score.}
    \label{fig:corner-cases}
    \vspace{-2mm}
\end{wrapfigure}
\noindent \textbf{Evaluation on Risk Detection on NuInstruct.}
To evaluate the performance of \model on risk scenario perception, we conduct experiments on the NuInstruct benchmark. As shown in Table~\ref{tab:nuinstruct_omnidrive} (Left), most MLLM-based models struggle with the risk detection task (MAP), revealing a fundamental limitation in their spatial reasoning capabilities. While InMLLM and VGGDrive improve performance by incorporating BEVFormer~\cite{li2024bevformer} and VGGT~\cite{wang2025vggt} into their inference pipelines, this introduces substantial computational overhead at test time. In contrast, our approach achieves superior performance across all metrics—including the lowest MAE (2.01), highest Accuracy (64.81), highest MAP (48.45), and highest BLEU (90.66)—without relying on any auxiliary 3D models at inference. These results demonstrate that distilling geometric priors during training enables \model to develop strong risk perception capabilities while preserving deployment efficiency.

\noindent \textbf{Evaluation on Language Caption on OmniDrive.}
%To further verify that injecting geometric constraints does not compromise the linguistic intelligence of the underlying MLLM, 
To verify that geometric constraints do not degrade the linguistic capability of the base MLLM
we evaluate \model on the OmniDrive benchmark, which focuses on caption-related tasks where base VLMs typically excel. As shown in Table~\ref{tab:nuinstruct_omnidrive} (Right), \model achieves the best overall performance among fine-tuned methods, with notable gains on CIDEr (96.70) and ROUGE (40.44), and a competitive BLEU score (37.88), yielding the highest average of 58.34. Compared to OmniDrive~\cite{wang2025omnidrive}, HERMES~\cite{zhou2025hermes}, and VGGDrive~\cite{wang2026vggdrive}, our method consistently delivers superior captioning quality, indicating that geometric distillation not only enhances spatial reasoning but also enriches the model's ability to describe complex driving scenes. This confirms that the spatial-aware design of \model complements rather than conflicts with the language generation capability inherited from the base MLLM.

\begin{table}[t]
\centering
\caption{\textbf{Left:} \textbf{Risk detection results on NuInstruct}. $\text{Avg.}^\ast = \max\left(\frac{\text{Accuracy} + \text{MAP} + \text{BLEU} - \text{MAE}}{4}, 0\right)$. \textbf{Right:} \textbf{Language caption on OmniDrive}. ZS and FT denote Zero-shot and Fine-tuned, respectively. $^\ddagger$ shows results take from \cite{wang2026vggdrive}.}
\vspace{-3mm}
\label{tab:nuinstruct_omnidrive}
\begin{minipage}[t]{0.49\textwidth}
    \centering
    \tabcolsep=0.1cm
    \renewcommand{\arraystretch}{1.05}
    \resizebox{\textwidth}{!}{%
    \begin{tabular}{l|lccccc}
        \toprule
        & Model & MAE$\downarrow$ & Acc.$\uparrow$ & MAP$\uparrow$ & BLEU$\uparrow$ & Avg.$^\ast\uparrow$ \\
        \midrule
        \multirow{4}{*}{\rotatebox{90}{ZS}}
          & GPT-4o~\cite{hurst2024gpt}$^\ddagger$        & 9.93  & 10.64 & 0     & 7.08  & 1.95  \\
          & LLAVA-OV~\cite{li2024llava}$^\ddagger$        & 87.04 & 3.75  & 0     & 8.55  & 0     \\
          & RoboTron~\cite{huang2025robotron}$^\ddagger$  & 19.36 & 2.57  & 0     & 8.06  & 0     \\
          & Qwen2.5-VL$^\ddagger$                         & 24.10 & 0.63  & 0     & 5.56  & 0     \\
        \midrule
        \multirow{4}{*}{\rotatebox{90}{FT}}
          % & Baseline                          & 4.35  & 47.71 & 6.15  & 75.75 & 31.32 \\
          & Baseline & 2.47 & 59.51 & 35.32 & 86.71 & 44.81 \\
          & InMLLM~\cite{ding2024holistic}$^\ddagger$     & 9.08  & 32.48 & 21.93 & 35.20 & 20.13 \\
          & VGGDrive~\cite{wang2026vggdrive}$^\ddagger$                            & 3.08  & 56.37 & 37.49 & 81.13 & 42.98 \\
          & \cellcolor{qwenblue} \textbf{\model}                   & \cellcolor{qwenblue}\textbf{2.01} & \cellcolor{qwenblue}\textbf{64.81} & \cellcolor{qwenblue}\textbf{40.36} & \cellcolor{qwenblue}\textbf{90.66} & \cellcolor{qwenblue}\textbf{48.45} \\
        \bottomrule
    \end{tabular}%
    }
\end{minipage}
\hfill
\begin{minipage}[t]{0.45\textwidth}
    \centering
    \tabcolsep=0.1cm
    \renewcommand{\arraystretch}{1.05}
    \resizebox{\textwidth}{!}{%
    \begin{tabular}{l|lcccc}
        \toprule
        & Model & BLEU$\uparrow$ & CIDEr$\uparrow$ & ROUGE$\uparrow$ & Avg.$\uparrow$ \\
        \midrule
        \multirow{3}{*}{\rotatebox{90}{ZS}}
          & GPT-4o~\cite{hurst2024gpt}$^\ddagger$         & 10.91 & 24.42 & 22.34 & 19.22 \\
          & LLAVA-OV~\cite{li2024llava}$^\ddagger$        & 16.14 & 28.41 & 22.14 & 22.23 \\
          & RoboTron~\cite{huang2025robotron}$^\ddagger$  & 20.30 & 34.33 & 23.67 & 26.10 \\
        \midrule
        \multirow{5}{*}{\rotatebox{90}{FT}}
          & Baseline                                       & 25.07 & 94.93 & 37.22 & 52.41 \\
          & OmniDrive~\cite{wang2025omnidrive}$^\ddagger$  & 38.00 & 68.60 & 32.60 & 46.40 \\
          & HERMES~\cite{zhou2025hermes}$^\ddagger$        & --    & 74.10 & 32.70 & --    \\
          & VGGDrive~\cite{wang2026vggdrive}$^\ddagger$    & 37.58 & 86.57 & 34.40 & 52.85 \\
          % \rowcolor{qwenblue}
          &  \cellcolor{qwenblue}\textbf{\model} & \cellcolor{qwenblue}\textbf{37.88} & \cellcolor{qwenblue}\textbf{96.70} & \cellcolor{qwenblue}\textbf{40.44} & \cellcolor{qwenblue}\textbf{58.34}  \\
        \bottomrule
    \end{tabular}%
    }
\end{minipage}
% \vspace{-8mm}
\end{table}

% \begin{wraptable}{r}{0.5\textwidth}
%     \centering
%     \tabcolsep=0.1cm
%     \renewcommand{\arraystretch}{1.05}
%     \vspace{-1.2em}
%     \caption{The performance comparison on the \textbf{NuInstruct} dataset \cite{ding2024holistic} against existing SOTA methods. $^\ast$ indicates $\max\left(\frac{\text{Accuracy} + \text{MAP} + \text{BLEU} - \text{MAE}}{4}, 0\right).$}
%     \label{tab-2}
%     \resizebox{0.5\textwidth}{!}{%
%     \begin{tabular}{l|lccccc}
%         \toprule
%         & Model & MAE$\downarrow$ & Acc.$\uparrow$ & MAP$\uparrow$ & BLEU$\uparrow$ & Avg.$^\ast\uparrow$ \\
%         \midrule
%         \multirow{4}{*}{\rotatebox{90}{Zero-shot}}
%           & GPT-4o \cite{hurst2024gpt}        & 9.93  & 10.64 & 0     & 7.08  & 1.95  \\
%           & LLAVA-OV \cite{li2024llava}       & 87.04 & 3.75  & 0     & 8.55  & 0     \\
%           & RoboTron \cite{huang2025robotron} & 19.36 & 2.57  & 0     & 8.06  & 0     \\
%           & Qwen2.5-VL                        & 24.10 & 0.63  & 0     & 5.56  & 0     \\
%         \midrule
%         \multirow{4}{*}{\rotatebox{90}{Fine-tuned}}
%           & Baseline                          & 4.35  & 47.71 & 6.15  & 75.75 & 31.32 \\
%           & InMLLM \cite{ding2024holistic}    & 9.08  & 32.48 & 21.93 & 35.2  & 20.13 \\
%           & VGGDrive                          & 3.08  & 56.37 & 37.49 & 81.13 & 42.98 \\
          
%           & \rowcolor{qwenblue} \textbf{\model}                   & \textbf{2.01} & \textbf{64.81} & \textbf{40.36} & \textbf{90.66} & \textbf{48.45} \\
%         \bottomrule
%     \end{tabular}%
%     }
%     \vspace{-1em}
% \end{wraptable}

\noindent \textbf{Evaluation on Generalization.} To further assess the robustness of our approach, we evaluate the generalization ability of \model under challenging scenarios by finetuning the baseline model (Qwen3VL-4B) on the DriveLM dataset~\cite{sima2024drivelm} and testing on DriveBench~\cite{xie2025vlms} (Figure~\ref{fig:corner-cases}). The results show that distilling knowledge from a powerful teacher not only improves the baseline's spatial intelligence but also enhances its generalization, with \model consistently achieving higher performance than the baseline across all scenarios.  
See Appendix~\ref{supp:qualitative} and the supplementary material for additional qualitative results.

\subsection{Ablation Study}

\begin{table}[h]
  \centering
  \begin{minipage}[t]{0.58\linewidth}
    \centering
    \setlength{\tabcolsep}{2pt}
    \caption{Effectiveness of each module on nuScenes.}
    \vspace{-2mm}
    \label{tab:ablation_modules}
    \resizebox{\linewidth}{!}{
    \begin{tabular}{l|ccc|ccc}
      \toprule
      \textbf{Exp.} & \textbf{\shortstack{Camera Pose\\Injector}} & \textbf{\shortstack{Multi-view\\EpipolarAttention}} & \textbf{\shortstack{3D Geometry\\Distillation}}  & \textbf{L2} $\downarrow$ & \textbf{Collision (\%)} $\downarrow$ & \textbf{Intersection (\%)} $\downarrow$ \\
      \midrule
      \#1 & \xmark        & \xmark         & \xmark         & 0.36 & 0.36 & 3.05 \\
      \#2 & \cmark & \xmark         & \xmark         &  0.32    &    0.24  & 2.80     \\
      \#3 & \xmark         & \xmark & \cmark         &  0.30    &    0.23  & 2.92     \\
      \#4 & \xmark         & \cmark       & \xmark &   0.33   &    0.23  &   2.84   \\
      \#5 & \cmark & \xmark & \cmark        & 0.29 & 0.21 & 2.52 \\
      \#6 & \cmark & \cmark & \cmark & \textbf{0.28} & \textbf{0.19} & \textbf{1.27} \\
      \bottomrule
    \end{tabular}
    }
  \end{minipage}
  \hfill
\begin{minipage}[t]{0.41\linewidth}
    \centering
    \setlength{\tabcolsep}{2pt}
    \caption{Temporal Sampling (N) on nuScenes. Time of feature extraction.}
    \vspace{-2mm}
    \label{tab:ablation_distillation}
    \resizebox{\linewidth}{!}{
    \begin{tabular}{c|cccc}
      \toprule
      \textbf{N} & \textbf{L2} $\downarrow$ & \textbf{Collision (\%)} $\downarrow$ & \textbf{Intersection (\%)} $\downarrow$ & \textbf{Time (h)} $\downarrow$ \\
      \midrule
      0 & 0.32 & 0.26 & 2.18 & 10 \\
      2 & 0.30 & 0.22 & 1.67 & 27 \\
      4 & \textbf{0.28} & \textbf{0.19} & \textbf{1.27} & \textbf{82} \\
      6 & 0.28 & 0.20 & 1.23 & 110 \\
      \bottomrule
    \end{tabular}
    }
\end{minipage}
  % \begin{minipage}[t]{0.40\linewidth}
  %   \centering
  %   \caption{Effectiveness of temporal sampling ($N$) in 3D geometric distillation. \textcolor{red}{Computational cost visualization}}
  %   \vspace{-2mm}
  %   \label{fig:temporal_sampling}
  %   \includegraphics[width=\linewidth]{images/line_chart.png}
  % \end{minipage}
  
\end{table}

\noindent \textbf{Component-wise Analysis.} Table~\ref{tab:ablation_modules} summarizes the contribution of each proposed module. Compared with the vanilla fine-tuned baseline (Exp.\#1), each component improves trajectory error and/or safety-related metrics.
%, showing that camera-pose conditioning, geometric distillation, and epipolar-guided feature alignment each contribute to spatial reasoning. 
The Camera Pose Injector (Exp.\#2) and 3D Geometry Distillation (Exp.\#3) provide strong individual gains, while Epipolar Feature Alignment (Exp.\#4) also improves all reported metrics by enforcing geometry-aware cross-view interaction. Combining pose conditioning with geometric distillation (Exp.\#5) further improves performance, and adding Epipolar Feature Alignment yields the best overall results. Together (Exp.\#6), these trends suggest that the three components address complementary aspects of spatial reasoning, including camera-pose grounding, 3D geometric supervision, and cross-view consistency.

\begin{table}[h]
  \centering
  \begin{minipage}[t]{0.45\linewidth}
    \centering
    \caption{Effectiveness of different ray representation on nuScenes open-loop planning.}
    \vspace{-2mm}
    \label{tab:ablation_pos_encoding}
    \resizebox{\linewidth}{!}{
    \begin{tabular}{c|ccc}
      \toprule
      \textbf{Ray} & \textbf{L2} $\downarrow$ & \textbf{Collision (\%)} $\downarrow$ & \textbf{Intersection (\%)} $\downarrow$ \\
      \midrule
      -                & 0.29  & 0.22 & 2.66 \\
      Na\"ive Ray      &   0.29   &  0.24    &    2.50  \\
      Pl\"ucker Ray    & \textbf{0.28} & \textbf{0.19} & \textbf{1.27} \\
      \bottomrule
    \end{tabular}
    }
  \end{minipage}
  \hfill
  \begin{minipage}[t]{0.49\linewidth}
    \centering
    \caption{Effectiveness of different dual attention mechanism on nuScenes open-loop planning.}
    \vspace{-2mm}
    \label{tab:ablation_attention}
    \resizebox{\linewidth}{!}{
    \begin{tabular}{c|ccc}
      \toprule
      \textbf{Attention} & \textbf{L2} $\downarrow$ & \textbf{Collision (\%)} $\downarrow$ & \textbf{Intersection (\%)} $\downarrow$ \\
      \midrule
      -                  & 0.29 & 0.21 & 2.52 \\
      Vanilla Attention  &  0.29    &   0.24   &    1.96  \\
      Epipolar Attention & \textbf{0.28} & \textbf{0.19} & \textbf{1.27} \\
      \bottomrule
    \end{tabular}
    }
  \end{minipage}
\end{table}

\noindent \textbf{Different Temporal Sampling Configuration.} Table~\ref{tab:ablation_distillation} studies the temporal sampling window $N$ used for 3D geometric distillation. The time column reports offline teacher feature-extraction time measured on a single NVIDIA A100 40GB GPU.
%Table~\ref{tab:ablation_distillation} studies the impact of the temporal sampling parameter $N$ in 3D Geometry Distillation. 
Increasing $N$ improves trajectory error and safety-related metrics over the single-frame setting, indicating that additional frames help the teacher recover more reliable 3D structure through multi-view consistency while increasing feature-extraction time. The gains saturate at $N=4$, while increasing to $N=6$ brings only marginal improvement at a higher distillation cost. We therefore adopt $N=4$ as the default setting, which provides sufficient temporal context for geometric supervision while avoiding unnecessary overhead.

%\noindent \textbf{Different Temporal Sampling Configuration.} Table~\ref{tab:ablation_distillation} investigates the impact of the temporal sampling parameter $N$ in 3D Geometry Distillation. With $N=0$, the model achieves 0.32 L2, 0.26\% Collision, and 2.18\% Intersection, indicating that single-frame geometry alone offers limited spatial cues. Increasing $N$ to 2 yields a clear improvement (0.30 L2, 1.67\% Intersection), as additional frames help the teacher recover more reliable 3D structure through multi-view consistency. Setting $N=4$ further reduces all metrics to 0.28 L2, 0.19\% Collision, and 1.27\% Intersection, while pushing $N$ to 6 brings only marginal changes (0.28 L2, 1.23\% Intersection) at substantially higher distillation cost. We therefore adopt $N=4$ as the default, providing sufficient temporal context for accurate geometric supervision without unnecessary overhead. 

\noindent \textbf{Different Pose Embedding Analysis.} Table~\ref{tab:ablation_pos_encoding} compares different positional embedding schemes within the Camera Pose Injector. Na\"ive Ray encoding, which uses only ray directions, provides limited improvement over the variant without explicit ray encoding, suggesting that direction alone is insufficient to disambiguate spatial relations across views. In contrast, Pl\"ucker Rays jointly encode ray direction and origin and achieve the best results across all metrics, with a particularly clear improvement in Intersection rate. This indicates that explicitly incorporating both viewing orientation and spatial offset provides a richer geometric prior for cross-view correspondence and scene-layout-aware trajectory prediction.

%\noindent \textbf{Different Pose Embbeding Analysis.} Table~\ref{tab:ablation_pos_encoding} compares different positional embedding schemes within the Camera Pose Injector. Without any explicit ray encoding, the model achieves 0.29 L2, 0.22\% Collision, and 2.66\% Intersection, reflecting limited awareness of per-pixel viewing geometry. Introducing a Naïve Ray embedding, which encodes only ray directions, yields negligible gains on L2 (0.29) and even a slight degradation in Collision (0.24\%), suggesting that direction alone is insufficient to disambiguate spatial relations across views. In contrast, the Plücker Ray representation which jointly encodes ray direction and origin, delivers the best performance (0.28 L2, 0.19\% Collision, 1.27\% Intersection), with a particularly sharp drop in Intersection rate. This indicates that explicitly encoding both the orientation and spatial offset of each ray provides a richer geometric prior, enabling the model to better reason about cross-view correspondences and produce trajectories that more faithfully respect scene layout.

\noindent \textbf{Different Attention Mechanism Analysis.} Table~\ref{tab:ablation_attention} compares attention designs for cross-view interaction. Vanilla Attention improves Intersection rate but does not consistently improve all metrics, suggesting that unconstrained global attention may introduce noisy cross-view correspondences. In contrast, Epipolar Attention restricts feature interaction to geometrically valid epipolar regions and achieves the best results across all metrics. This confirms that geometry-constrained attention provides more reliable cross-view alignment for spatially consistent trajectory prediction.

%\noindent \textbf{Different Attention Mechanism Analysis.} Table~\ref{tab:ablation_attention} examines different attention designs for cross-view interaction. Without any dedicated attention mechanism, the model attains 0.29 L2, 0.21\% Collision, and 2.52\% Intersection, leaving clear room for cross-view feature refinement. Replacing it with Vanilla Attention provides only marginal gains (0.29 L2, 0.24\% Collision, 1.96\% Intersection), suggesting that unconstrained global attention spreads correspondences too broadly across views and dilutes the geometric signal. In contrast, Epipolar Attention—which restricts cross-view interaction along geometrically valid epipolar lines—achieves the best performance (0.28 L2, 0.19\% Collision, 1.27\% Intersection), demonstrating that constraining attention along epipolar geometry better captures cross-view correspondences and yields more spatially consistent trajectory predictions.

% \subsection{Qualitative Results}

\section{Related Works}
% \textcolor{red}{HAO: I will revise this section}

\noindent
\textbf{Geometry-Aware in MLLMs}
A growing body of work observes that MLLMs trained under image--text supervision lack the \textit{Spatial Intelligence} required to understand 3D scenes~\cite{yang2025thinking, yang2025cambrian, zhang2026theory}, as they struggle to construct internal representations of scene geometry. To address this, recent methods adopt straightforward recipes built on 3D foundation models~\cite{wang2024dust3r, wang2025vggt}: Spatial-MLLM~\cite{wu2025spatial} and VG-LLM~\cite{zheng2025learning} simply fuse 2D features with 3D features at the input stage, while 3DRS~\cite{huang20253drs} and 3DThinker~\cite{chen2025think} merely apply a feature-distillation loss at the MLLM's last layer. This lightweight paradigm has been validated almost exclusively on indoor, static, or object-centric scenes, where short baselines and high inter-view overlap let off-the-shelf 3D teachers produce clean supervision that a single alignment loss can transfer. Driving scenes invert these properties: surround views with limited overlap, large depth ranges, and pervasive motion cause the same teachers to yield noisy, incomplete signals (Figure~\ref{fig:vggt_issue}). Existing realizations thus do not transfer to driving without rethinking how supervision is generated and consumed.

\noindent
\textbf{MLLMs for Autonomous Vehicles}
Recent advances in MLLMs~\cite{bai2025qwen3,wang2025internvl3,team2024gemma,yao2024minicpm} have driven a wave of driving systems, spanning language-conditioned models such as DriveGPT4~\cite{xu2024drivegpt4}, DriveLM~\cite{sima2024drivelm}, and DriveVLM~\cite{tian2025drivevlm}, as well as unified frameworks including EMMA~\cite{hwang2024emma}, ORION~\cite{fu2025orion}, OmniDrive~\cite{wang2025omnidrive}, and RoboTron-Drive~\cite{huang2025robotron}. As our diagnostic analysis in Section~\ref{sec:investigate} shows, these models treat AV tasks as a 2D autoregressive problem and consequently struggle with depth and cross-view correspondence. The closest works to ours mitigate this by coupling the VLM with an external 3D model at inference: VGGDrive~\cite{wang2026vggdrive} routes VGGT~\cite{wang2025vggt} features through hierarchical cross-attention, and SpaceDrive~\cite{li2025spacedrive} lifts visual patches into a common geometric space via a frozen depth estimator. Both depend on auxiliary 3D modules at \emph{inference time}, which as reported in Table~\ref{tab:open_loop}, drops throughput to $0.03$ and $0.25$\,FPS respectively and exposes predictions to cascading failures of the upstream estimator. OmniSpace departs from this paradigm by confining 3D supervision to training, internalizing geometric awareness within the MLLM itself and avoiding any test-time 3D dependency.

 \section{Conclusion \& Discussion}
 % \nghicomment{[Nghi]: Temporal averaging assumes teacher features are aligned across time. Please explain whether ego-motion compensation or world-frame alignment is used, and how moving objects are handled.}
We presented \textbf{OmniSpace}, a geometry-aware adaptation framework that strengthens depth estimation and cross-view correspondence directly within MLLMs, without requiring an external 3D model at inference. Through Camera Pose Injection, Multi-view Epipolar Attention, and 3D Geometric Distillation with temporal aggregation, OmniSpace internalizes geometric priors at training time and consistently improves planning, risk detection, language, and generalization performance across five autonomous-driving benchmarks while preserving the inference efficiency of pure 2D MLLMs.

\noindent
\textbf{Discussion.} Several aspects of OmniSpace leave room for further refinement. First, our temporal aggregation averages VGGT features over a symmetric window and could be made more precise by accounting for ego-motion when fusing neighboring frames. Second, the Plücker-ray and epipolar modules assume reasonably accurate camera calibration, which holds in standard AV datasets but may need additional robustness handling under sensor drift or recalibration. Finally, we adopt a single fixed teacher (VGGT); exploring stronger or AV-pretrained 3D foundation models is a straightforward direction we leave for future work.

% \clearpage
{\small
\bibliographystyle{plain}
\bibliography{main}
}

\clearpage
\appendix
%  Appendix-only PDF link style (main body keeps default hyperref appearance). 
\hypersetup{%
  colorlinks=true,%
  linkcolor=appendixlink,%
  citecolor=appendixlink,%
  urlcolor=appendixlink,%
  filecolor=appendixlink,%
  menucolor=appendixlink,%
  runcolor=appendixlink,%
  pdfborder={0 0 0}%
}

\renewcommand{\thepart}{}%
\renewcommand{\partname}{}%
\part{Appendix} %
\parttoc %
\clearpage

\newpage
\clearpage
\renewcommand{\thetable}{S-\Roman{table}}
\renewcommand{\thefigure}{S-\Roman{figure}}

\renewcommand{\thesection}{\Alph{section}}
\renewcommand{\theHsection}{supp.\Alph{section}}

\setcounter{section}{0}
\setcounter{table}{0}
\setcounter{figure}{0}
\setcounter{page}{0}

The supplementary material provides additional details on data construction, evaluation protocols, additional analysis and qualitative results to further support our main findings: \\

 \textbf{A. Geo-Bench Data Construction}: We describe how Geo-Bench, the diagnostic benchmark used in Sec.~\ref{sec:investigate}, is constructed from five autonomous-driving datasets, including the coordinate unification procedure and the QA generation pipeline for evaluating depth understanding and cross-view correspondence.

\textbf{B. Detailed Datasets and Metrics}: We provide an in-depth description of the benchmarks used to evaluate \model, covering nuScenes, Bench2Drive, NuInstruct, OmniDrive, and DriveBench, along with their task setups and evaluation metrics.

\textbf{C. Further Analysis}: We further report the performance of OmniSpace under calibration noise and analyze geometry awareness through and comparisons with other Geometry-Aware AV MLLMs on Geo-Bench.

 \textbf{D. Additional Qualitative Results}: We present further qualitative examples across NuInstruct, OmniDrive, nuScenes open-loop planning, and Bench2Drive closed-loop planning, with additional video results provided in the supplementary material.

\section{Geo-Bench Data Construction}
\label{supp:data_construction}
This appendix describes how Geo-Bench, the diagnostic benchmark used in Section~\ref{sec:investigate}, is constructed. Geo-Bench is built from the validation splits of five autonomous-driving datasets (nuScenes~\cite{caesar2020nuscenes}, Argoverse~2~\cite{wilson2023argoverse}, Waymo~\cite{sun2020scalability}, ONCE~\cite{mao2021one}, and TruckScenes~\cite{fent2024man}) and is designed to isolate two complementary geometric abilities of MLLMs: \textit{depth understanding}, evaluated through QA pairs, and \textit{cross-view correspondence}, evaluated through a representation-level probe. We describe the data preparation, the QA generation, and the probing protocol below.

\subsection{Data Preparation and Coordinate Unification}
\label{supp:coordinate_unification}
The five source datasets do not share a common annotation format. They differ in two key aspects: (i) the reference frame in which 3D bounding boxes are expressed, and (ii) the convention used for the yaw angle. To enable a consistent multi-view geometric evaluation, we unify all annotations into the nuScenes coordinate convention. For each non-nuScenes dataset, we apply a planar rotation $R$ around the $z$-axis with offset $\alpha$ to every bounding-box center $\mathbf{c}_i$ and yaw $\theta_i$:
\begin{align}
\mathrm{R} =
\begin{bmatrix}
\cos\alpha & -\sin\alpha & 0 \\
\sin\alpha & \cos\alpha  & 0 \\
0 & 0 & 1
\end{bmatrix}, \quad
\theta_i' = \theta_i + \alpha, \quad \mathbf{c}_i' = \mathrm{R}\,\mathbf{c}_i.
\end{align}
The angular offset $\alpha$ is set to $\pi/2$, $\pi/2$, $\pi$, and $3\pi/4$ for Argoverse~2, Waymo, ONCE, and TruckScenes, respectively. We rotate the camera extrinsic and ego-pose matrices accordingly so that all subsequent cross-view projections remain consistent. After unification, Geo-Bench draws from a shared pool of multi-view frames with consistent calibration, ego pose, and 3D box annotations.

\subsection{QA Construction}
\label{supp:depth_qa}
The depth-understanding subset contains 250 multi-view distance-estimation questions covering two complementary settings: \textit{ego-to-object}, where the model estimates the distance from the ego vehicle to a single referenced object, and \textit{object-to-object}, where the model estimates the distance between two referenced objects observed across multi-view images. Each question shows the model the surround-view images of a frame, with referenced objects highlighted by colored 2D bounding-box overlays drawn directly on the camera images: a \textbf{blue} box labeled \texttt{(1)} for the first (or only) referenced object and a \textbf{magenta} box labeled \texttt{(2)} for the second referenced object. Ground-truth distances are computed directly from the unified 3D bounding boxes as $d_i^{\text{ego}} = \|\mathbf{c}_i^{\text{ego}}\|$ for the ego-to-object setting and $d_{ij} = \|\mathbf{c}_i - \mathbf{c}_j\|$ for the object-to-object setting.

\paragraph{Object sampling.} For each scene, we restrict candidate objects to those (i) annotated as a vehicle, pedestrian, or cyclist, (ii) lying within $[3, 60]$\,m of the ego vehicle, and (iii) projected onto a single camera with at least $60\%$ of their 2D bounding box visible. For object-to-object questions, we additionally require the two objects to appear in different camera views so that the answer cannot be inferred from a single image, and we require the inter-object distance to lie in $[5, 50]$\,m. Objects are sampled to cover a broad range of distances rather than concentrating in any single regime, and we sample 50 questions per source dataset, balanced between the two question types.

\paragraph{Prompt templates.} The prompts are identical across all evaluated MLLMs. Multi-view inputs are presented as an ordered sequence of camera-name tags interleaved with their corresponding images. The number of views varies across datasets (e.g., six for nuScenes and Argoverse~2, five for Waymo, seven for TruckScenes); we show one representative six-camera example below.

\begin{center}
\fbox{\begin{minipage}{0.95\linewidth}
\small
\textbf{[System]} You are a helpful assistant for autonomous driving. You will be given a question and a list of images, and you need to answer the question based on the images. Please first do your thinking process in \texttt{<think> </think>} and then provide the final answer in \texttt{<answer> </answer>} tags. \\[4pt]
\textbf{[User]} \\
\texttt{<CAM\_NAME\_1>:} \texttt{<image>} \, \texttt{<CAM\_NAME\_2>:} \texttt{<image>} \, \ldots \, \texttt{<CAM\_NAME\_K>:} \texttt{<image>} \\[4pt]
\textit{(Ego-to-Object)} Given the current driving images, estimate the approximate distance from the marked object (highlighted with a blue bounding box) to our vehicle (ego vehicle / camera origin). \\[2pt]
\textit{(Object-to-Object)} Given the current driving images, estimate the approximate distance between Object-1 (marked with blue bounding box and label ``(1)'') and Object-2 (marked with magenta bounding box and label ``(2)'').\\[2pt]
Provide your answer as a single numerical value in meters (e.g., 15.5). \\[6pt]
\textit{Example camera tags for six-view datasets:} \\
\texttt{CAM\_FRONT\_LEFT}, \texttt{CAM\_FRONT}, \texttt{CAM\_FRONT\_RIGHT}, \texttt{CAM\_BACK\_LEFT}, \texttt{CAM\_BACK}, \texttt{CAM\_BACK\_RIGHT}.
\end{minipage}}
\end{center}

\section{Deatiled of Datasets and Metrics}
\label{supp:dataset_and_metric}

To demonstrate the effectiveness of \model in injecting geometric awareness into MLLMs, we evaluate \model across various attributes within the autonomous driving domain. These benchmarks cover task-specific scenarios, including open-loop planning, closed-loop planning, scene understanding, perception, cross-view risk detection, and generalization.

\noindent \textbf{nuScenes~\cite{caesar2020nuscenes}.} The nuScenes dataset comprises 1,000 urban driving scenes with full-stack 360$^{\circ}$ sensing (6 cameras, 1 LiDAR, and 5 radars). For open-loop planning, we predict 6 waypoints within a 3\,s horizon and evaluate: (i) \textit{waypoint displacement} (L2 error), (ii) \textit{collision rate} (the fraction of future timestamps overlapping with any dynamic agent), and (iii) \textit{intersection rate} (the fraction of timestamps intruding into non-drivable map regions).

\noindent \textbf{Bench2Drive~\cite{jia2024bench2drive}.} Bench2Drive is a closed-loop planning benchmark emphasizing interactive scenarios (e.g., merging, overtaking, yielding, and emergency negotiation) in a deterministic CARLA V2~\cite{dosovitskiy2017carla} simulator. Our closed-loop evaluation adopts the official protocol of 220 short routes, covering 44 interactive scenarios, with 5 distinct routes defined for each scenario. Closed-loop metrics include \textit{Driving Score} (route progress penalized by safety infractions) and \textit{Success Rate} (the percentage of scenarios completed without terminal violation). All reported results adopt identical horizon, temporal sampling, footprint inflation, and map definitions for fair comparison.

\noindent \textbf{NuInstruct~\cite{ding2024holistic}.} NuInstruct samples a total of 11,850 keyframes from 850 videos within the nuScenes dataset. It includes a wide range of challenging samples, such as cross-view risk perception, target distance estimation, agent and ego state prediction, target motion prediction, and reasoning tasks. The dataset effectively reflects a model's ability to perform cross-view understanding and analyze 3D geometric scene information. NuInstruct employs a variety of metrics to evaluate different tasks, including \textit{Mean Absolute Error} (MAE) for regression tasks, \textit{accuracy} for classification tasks, \textit{mean Average Precision} (mAP) for detection tasks, and a rule-based \textit{BLEU} metric for captioning tasks.

\noindent \textbf{OmniDrive~\cite{wang2025omnidrive}.} OmniDrive is also built upon the nuScenes dataset and primarily focuses on scene description, attention-based question answering, counterfactual reasoning question answering, decision-planning question answering, and general dialogue. Due to limitations in the availability of evaluation data and metrics, we follow~\cite{wang2026vggdrive} select scene description and general dialogue from this dataset to train and evaluate the model's ability to understand autonomous driving scene captions. The dataset primarily includes three language evaluation metrics: \textit{BLEU}, \textit{CIDEr}, and \textit{ROUGE}, with the ``average'' representing the mean of the three metrics.

\noindent \textbf{DriveBench~\cite{xie2025vlms}.} DriveBench is a benchmark dataset designed to assess VLM reliability across 17 settings, including clean, corrupted, and text-only inputs. It comprises 19,200 frames, 20,498 question--answer pairs, three question types, and four mainstream driving tasks (\textit{perception}, \textit{prediction}, \textit{planning}, and \textit{behavior explanation}), and evaluates 12 popular VLMs. Built upon DriveLM~\cite{sima2024drivelm}, DriveBench employs GPT-3.5-turbo as the judge model to score open-ended responses, and we follow the same protocol when evaluating our model.

% \section{Extended Ablation Studies}
% The main paper reports the currently completed ablations for module composition and attention design. Additional teacher-model and temporal-window ablations will follow the same nuScenes metrics once all runs are complete.

\section{Further Analysis}
\begin{table}[h]
  \centering

  \begin{minipage}[t]{0.48\linewidth}
    \centering
    \setlength{\tabcolsep}{2pt}
    \caption{Effect of calibration noise on performance.}
    \vspace{-2mm}
    \label{tab:calibration_noise}
    \resizebox{\linewidth}{!}{
    \begin{tabular}{l|ccc}
      \toprule
      \textbf{Calibration Noise}
      & \textbf{L2} $\downarrow$
      & \textbf{Collision (\%)} $\downarrow$
      & \textbf{Intersection (\%)} $\downarrow$ \\
      \midrule
      $\pm 10^\circ$; $\pm 10$ cm & 0.30 & 0.22 & 2.12 \\
      $\pm 5^\circ$; $\pm 5$ cm   & 0.29 & 0.21 & 1.28 \\
      $\pm 0^\circ$; $\pm 0$ cm   & \textbf{0.28} & \textbf{0.19} & \textbf{1.27} \\
      \bottomrule
    \end{tabular}
    }
  \end{minipage}
  \hfill
  \begin{minipage}[t]{0.51\linewidth}
    \centering
    \setlength{\tabcolsep}{2pt}
    \caption{Geometry-awareness comparison on Geo-Bench.}
    \vspace{-2mm}
    \label{tab:geometry_awareness}
    \resizebox{\linewidth}{!}{
    \begin{tabular}{l|cc}
      \toprule
      \textbf{Model}
      & \textbf{\shortstack{Cross-View\\Correspondence (\%)}} $\uparrow$
      & \textbf{\shortstack{Depth\\Estimation (m)}} $\downarrow$ \\
      \midrule
      VGGDrive                  & 48.65         & 43.36 \\
      SpaceDrive                & 27.25         & 54.36 \\
      \textbf{OmniSpace (ours)} & \textbf{59.8} & \textbf{14.4} \\
      \bottomrule
    \end{tabular}
    }
  \end{minipage}

\end{table}
\subsection*{Robustness of under calibration noise}
Since OmniSpace relies on calibrated camera geometry for Multi-view Epipolar Attention, we further evaluate its sensitivity to calibration errors by perturbing both the rotational and translational components of the camera calibration. As shown in Table~\ref{tab:calibration_noise}, performance degrades only moderately as the perturbation magnitude increases. Under $\pm5^\circ$ rotation and $\pm5$ cm translation noise, the L2 error increases from 0.28 to 0.29, while the collision and intersection rates change only slightly. Even under the stronger $\pm10^\circ$ and $\pm10$ cm perturbation, the model maintains stable performance. This robustness is partly attributed to the dilated epipolar region used in our attention mechanism, which helps absorb small calibration inaccuracies. These results indicate that OmniSpace is reasonably robust to moderate calibration noise and sensor drift of similar magnitude.

\subsection*{Comparison with Geometry-Aware AV MLLMs}
We further compare OmniSpace with existing geometry-aware AV MLLMs on Geo-Bench by reproducing VGGDrive~\cite{wang2026vggdrive} and SpaceDrive~\cite{li2025spacedrive} under the same evaluation protocol used for OmniSpace. Since the training code for VGGDrive is not publicly available and SpaceDrive is originally designed only for trajectory estimation, we adapt both methods to Geo-Bench and fine-tune them on DriveLM for 800 steps using the same setup as OmniSpace. As shown in Table~\ref{tab:geometry_awareness}, OmniSpace achieves 59.8\% cross-view correspondence accuracy and a depth estimation error of 14.4\,m, outperforming VGGDrive (48.65\%, 43.36\,m) and SpaceDrive (27.25\%, 54.36\,m). These results provide direct evidence that OmniSpace improves the two geometric capabilities targeted by our design, namely cross-view correspondence and depth estimation, which is consistent with the downstream AV performance observed in our experiments.

\section{Additional Qualitative Results}
\label{supp:qualitative}
\begin{figure}
    \centering
    \includegraphics[width=1.0\linewidth]{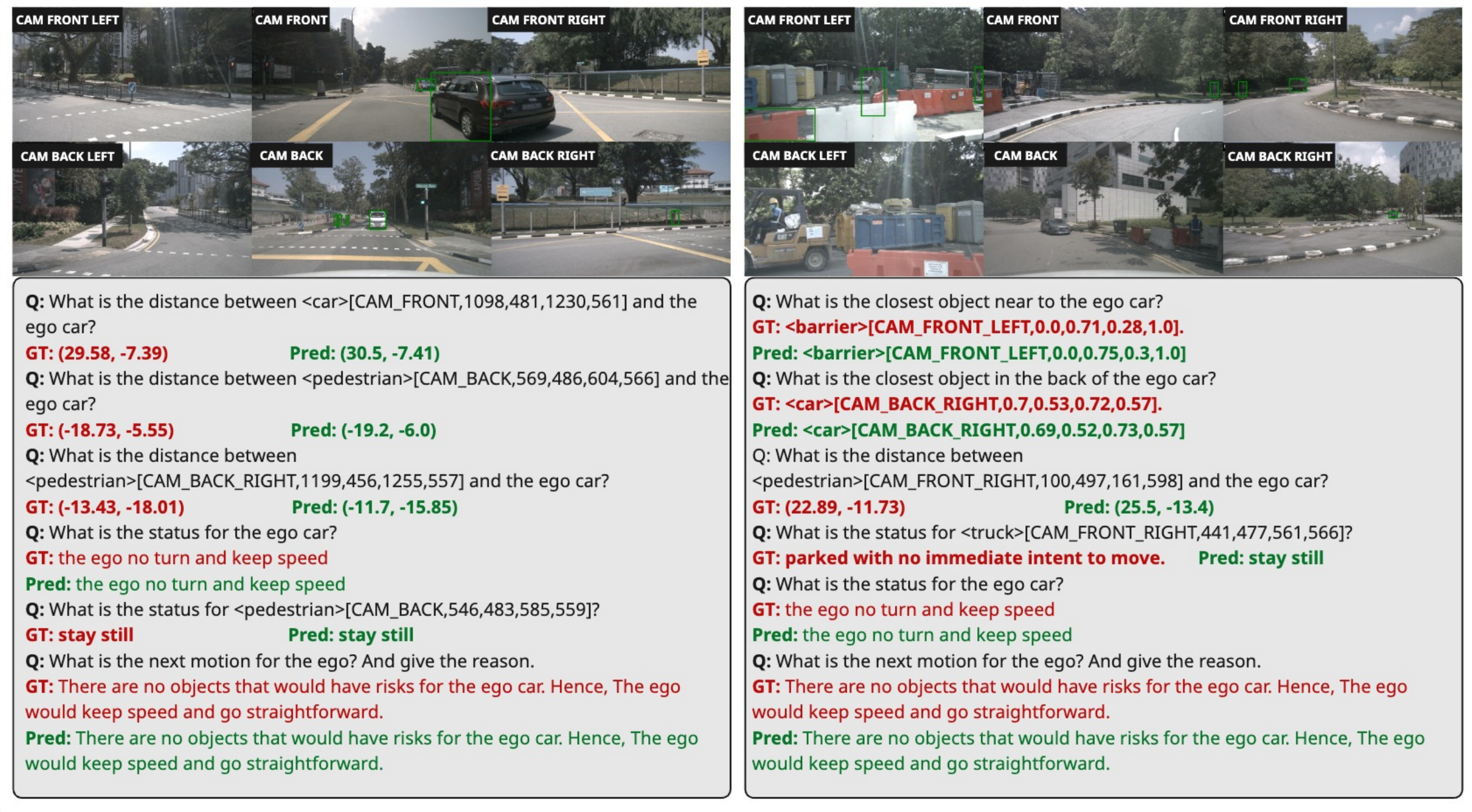}
    \caption{OmniSpace's results on NuInstruct compared to ground truths}
    \label{fig:nuinstructsupp}
\end{figure}

\begin{figure}
    \centering
    \includegraphics[width=1.0\linewidth]{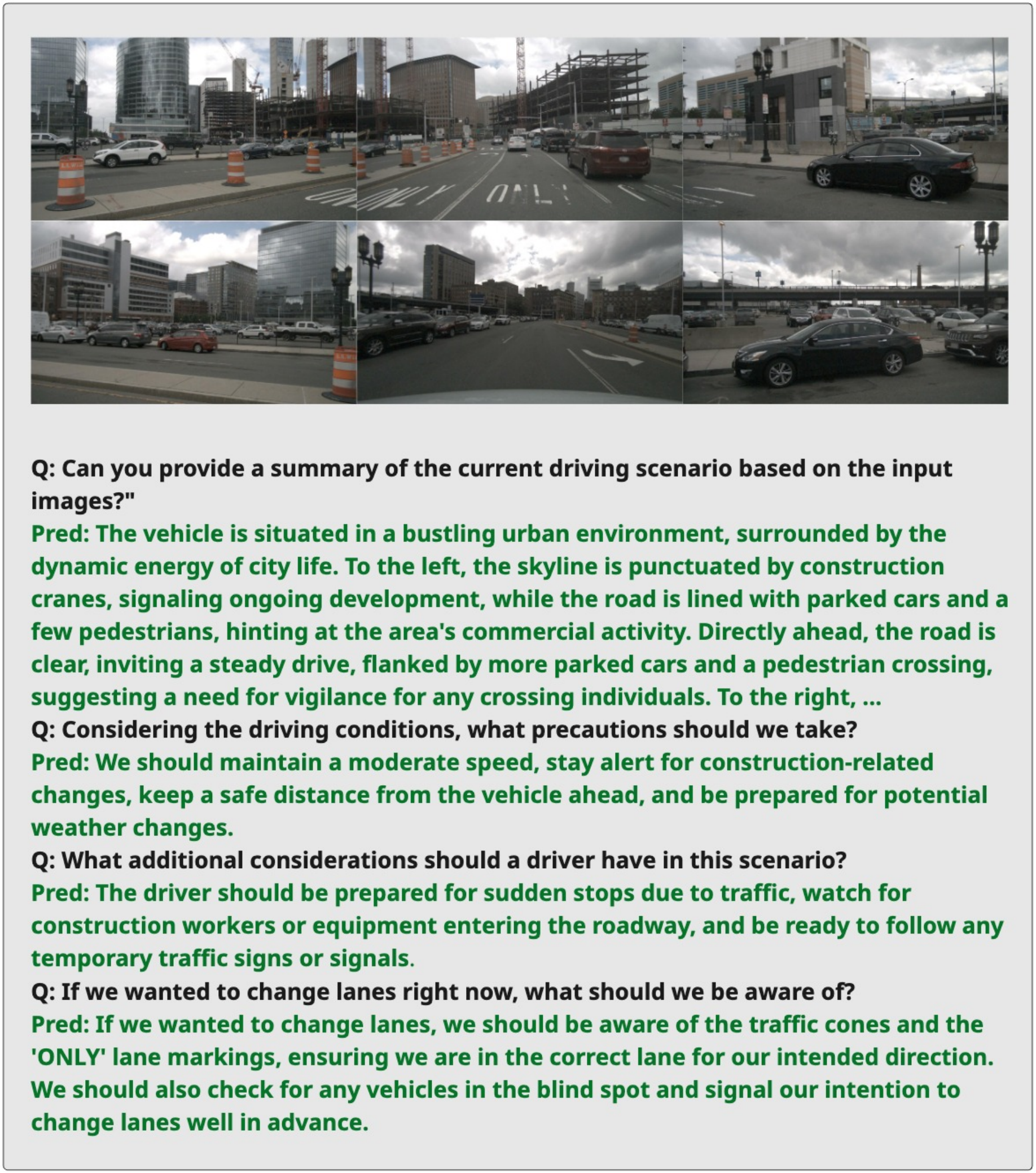} 
    \caption{An illustration of counterfactual reasoning ability of our method on OmniDrive dataset}
    \label{fig:omnisupp}
\end{figure}

\begin{figure}
    \centering
    \includegraphics[width=1.0\linewidth]{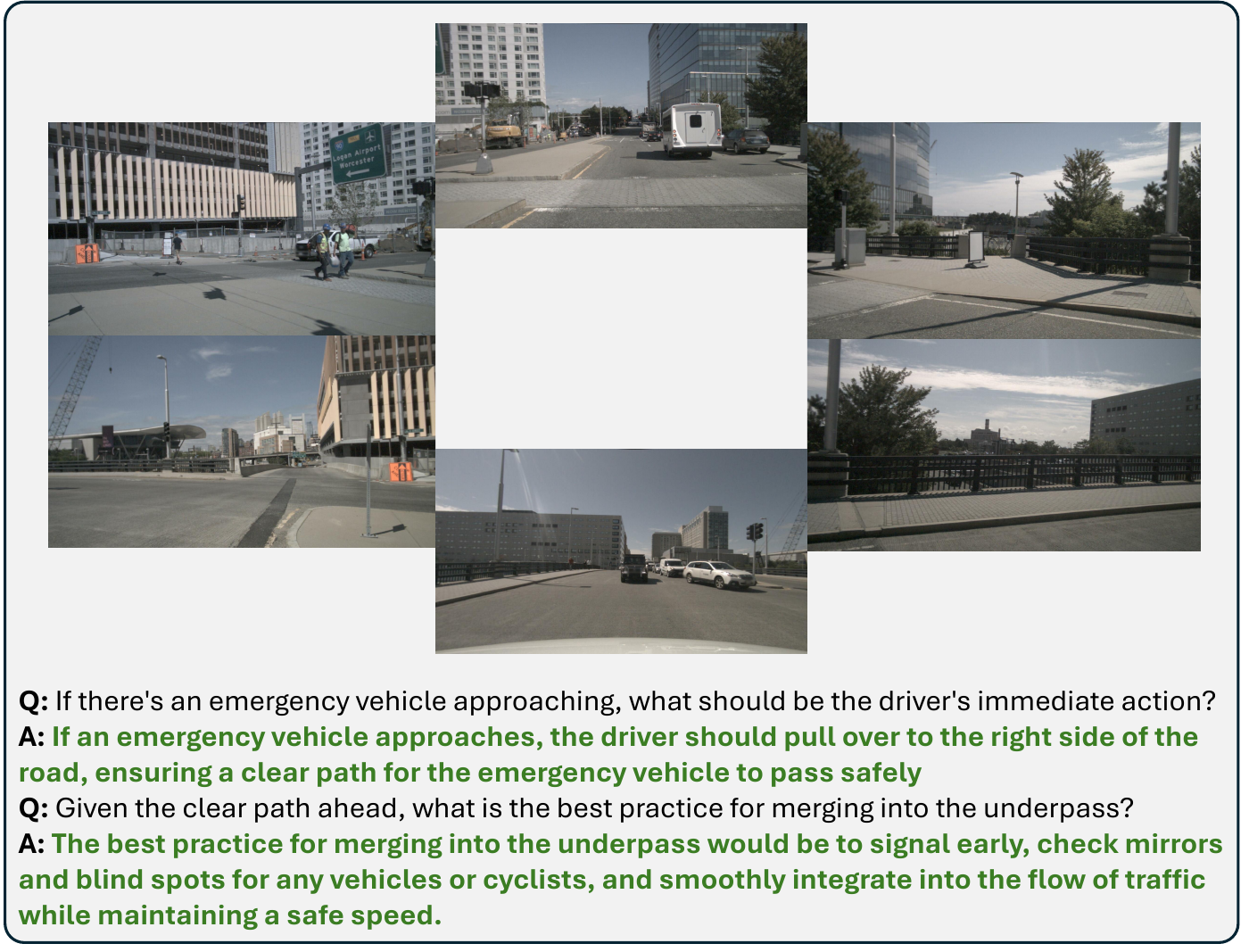}
    \caption{Another examples of counterfactual reasoning of our proposed solution}
    \label{fig:omnisupp2}
\end{figure}

\begin{figure}
    \centering
    \includegraphics[width=1.0\linewidth]{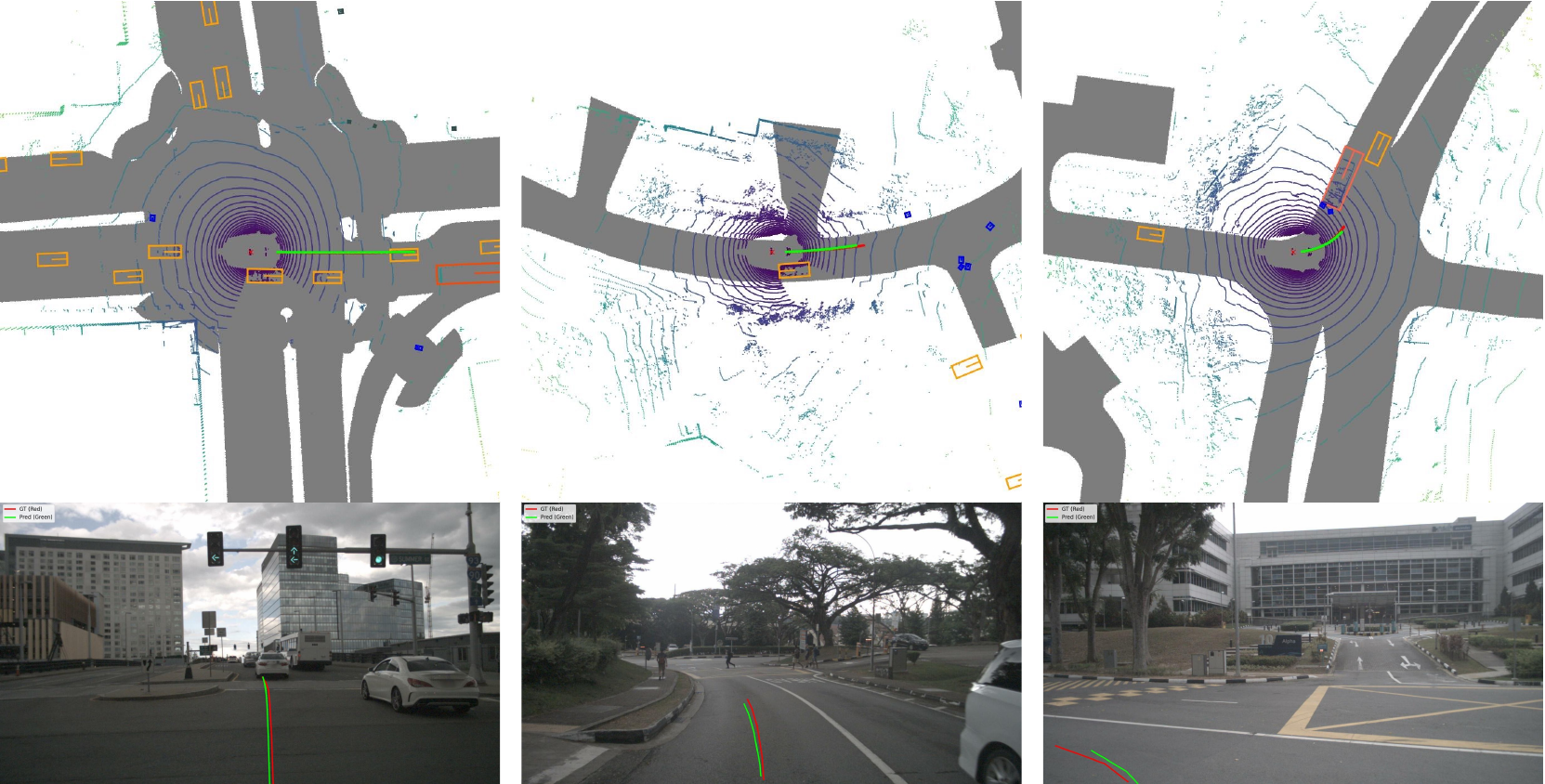}
    \caption{Qualitative results of our OmniSpace on planning task in nuScenes dataset}
    \label{fig:supp_openloop}
\end{figure}

\begin{figure}
    \centering
    \includegraphics[width=1.0\linewidth]{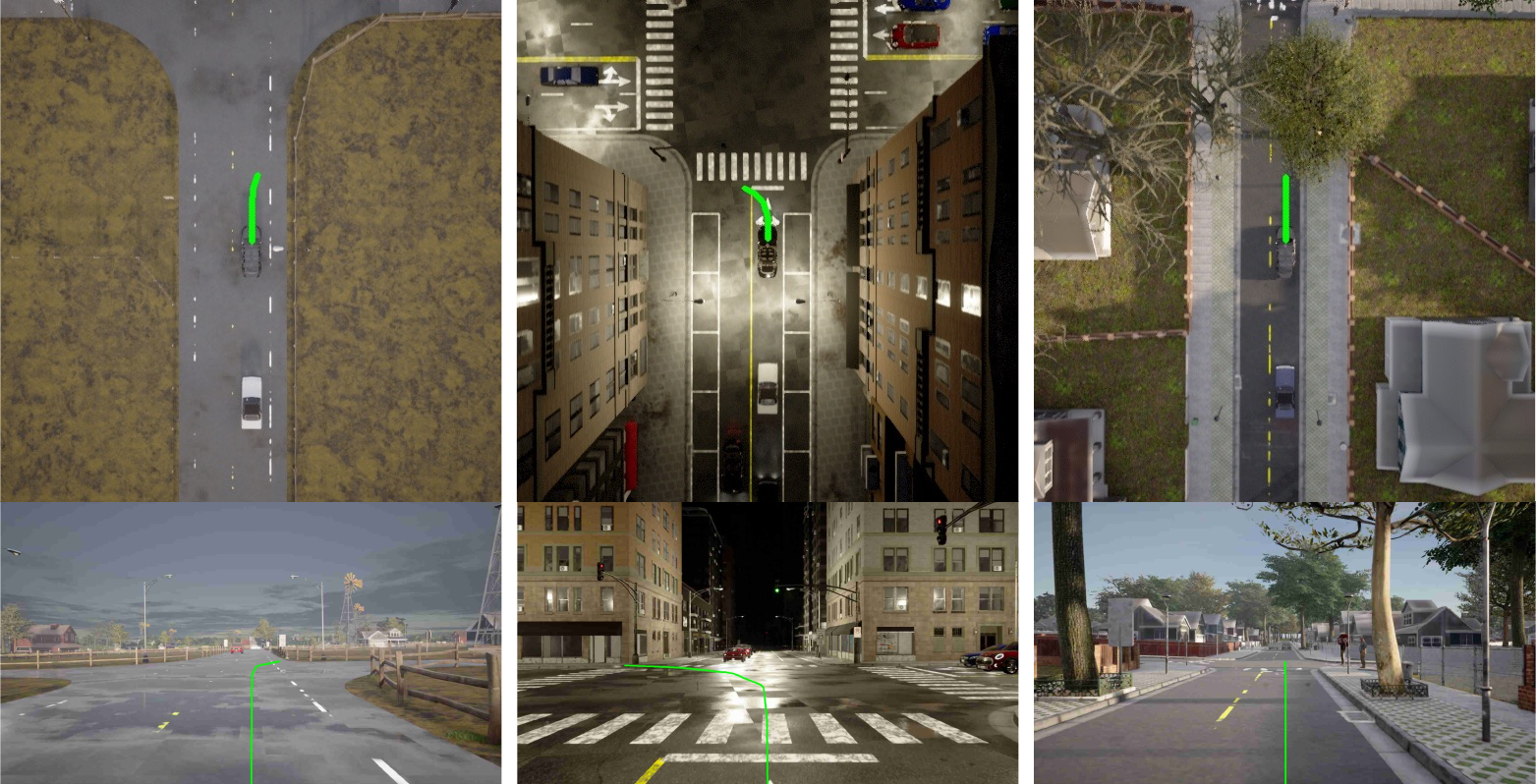}
    \caption{Qualitative results of our OmniSpace on planning tasks in the Bench2Drive dataset.}
    \label{fig:bench2drive_supp}
\end{figure}

Additional qualitative results are shown in Figure~\ref{fig:nuinstructsupp} for NuInstruct, Figures~\ref{fig:omnisupp} and~\ref{fig:omnisupp2} for OmniDrive, Figure~\ref{fig:supp_openloop} for nuScenes Open-Loop Planning, and Figure~\ref{fig:bench2drive_supp} for Bench2Drive. Additional qualitative video results for planning tasks are also provided in the supplementary material.
% Qualitative examples are used to inspect failure modes in object distance estimation, cross-view association, and planning outputs under challenging camera overlap and occlusion cases.

% \section{Additional Discussion}
% The proposed design improves geometry awareness without test-time 3D teachers, but it still depends on camera calibration quality and on the reliability of training-time geometric supervision.

\clearpage

% \newpage
% \input{checklist.tex}
\end{document}